\documentclass{article} 
\usepackage{iclr2025_conference,times}

\usepackage{amsmath,amsfonts,bm}

\def\eqref#1{equation~\ref{#1}}

\def\1{\bm{1}}

\DeclareMathAlphabet{\mathsfit}{\encodingdefault}{\sfdefault}{m}{sl}
\SetMathAlphabet{\mathsfit}{bold}{\encodingdefault}{\sfdefault}{bx}{n}

\usepackage{hyperref}
\usepackage{url}
\usepackage{graphicx}
\usepackage{inconsolata}
\usepackage{graphicx}
\usepackage{xcolor}
\definecolor{darkgreen}{rgb}{0.0, 0.5, 0.0}
\usepackage{colortbl}
\usepackage{inconsolata}
\usepackage{color}
\usepackage{graphicx}
\usepackage{amsmath}
\usepackage{amsfonts}
\usepackage{amsthm}
\usepackage{multirow}
\usepackage{array}
\usepackage{color}
\usepackage{makecell}
\usepackage{etoolbox}
\usepackage{url}

\usepackage{inconsolata}

\usepackage{graphicx}
\usepackage{amsmath}
\usepackage{amsfonts}
\usepackage{amsthm}
\usepackage{multirow}
\usepackage{array}
\usepackage{color}
\usepackage{makecell}
\usepackage{etoolbox}
\usepackage{url}
\usepackage{enumitem}
\usepackage{booktabs}
\usepackage{dirtytalk}
\usepackage{booktabs}
\usepackage{xcolor}
\definecolor{darkgreen}{RGB}{0, 100, 0}
\usepackage{pifont}
\usepackage{wrapfig}
\usepackage{caption}
\usepackage{alltt}
\usepackage{setspace}
\usepackage{float}

\title{SkillMentor: LLM Agent Self-Evolution via Learning Blind-Spot Diagnosis}

\author{
\textbf{Xiaoyi Bao\textsuperscript{1}\thanks{Work done during an internship at Tencent.}} \quad
\textbf{Yuanzhen Xie\textsuperscript{2}} \quad
\textbf{Yunzhi Tan\textsuperscript{2}$^\dagger$} \quad
\textbf{Jinghang Gu\textsuperscript{1}} \\
\textbf{Zhongqing Wang\textsuperscript{3}} \quad
\textbf{Chu-Ren Huang\textsuperscript{1}} \quad
\textbf{Bo Hu\textsuperscript{2}$^\dagger$} \quad
\textbf{Zang Li\textsuperscript{2}} \\[1ex]
\textsuperscript{1}The Hong Kong Polytechnic University \quad
\textsuperscript{2}Platform and Content Group, Tencent \\
\textsuperscript{3}Soochow University \\[0.5ex]
\texttt{p2213545413@outlook.com}, \;
\texttt{xieyzh3@gmail.com}, \;
\texttt{wangzq@suda.edu.cn} \\
\texttt{\{boristan,harryyfhu,gavinzli\}@tencent.com}\\
\texttt{\{jinghang.gu, churen.huang\}@polyu.edu.hk}
}

\definecolor{graytext}{RGB}{150, 150, 150}
\definecolor{diagpink}{RGB}{255, 220, 230}
\definecolor{transgreen}{RGB}{180, 235, 180}
\definecolor{transyellow}{RGB}{255, 245, 170}
\definecolor{gold}{RGB}{255, 180, 0}
\newcommand{\std}[1]{\textsubscript{\textit{\textcolor{graytext}{\scriptsize $\!#1\!$}}}}

\iclrfinalcopy 
\begin{document}

\maketitle
{\renewcommand\thefootnote{$\dagger$}\footnotetext{Corresponding author.}}

\begin{abstract}

Agent self-evolution has primarily focused on learning how to act, while overlooking an equally important capability: learning to discover what an agent does not know. Existing approaches typically assume that failure discovery is given, focusing on how to repair failures once they are identified. We ask whether blind-spot diagnosis itself can be learned.
We thus study diagnosis as an agent capability separate from execution, and exclude two alternative sources of progress: executor adaptation and human supervision. Under these constraints, performance cannot improve through executor updates or annotated examples, forcing all improvements to originate from the learned diagnostic capability.
We propose SkillMentor, which trains a Mentor policy via reinforcement learning to generate diagnostic tasks, identify recurrent failure modes, and curate them into reusable corrective skills. Across AppWorld and BFCLv3, SkillMentor improves executor performance by an average of 44.2\%. These results suggest that blind-spot diagnosis is a learnable capability, enabling self-evolution without updating executor weights or relying on human-curated data.

\end{abstract}

\begin{figure}[h]
\centering
\includegraphics[width=\linewidth]{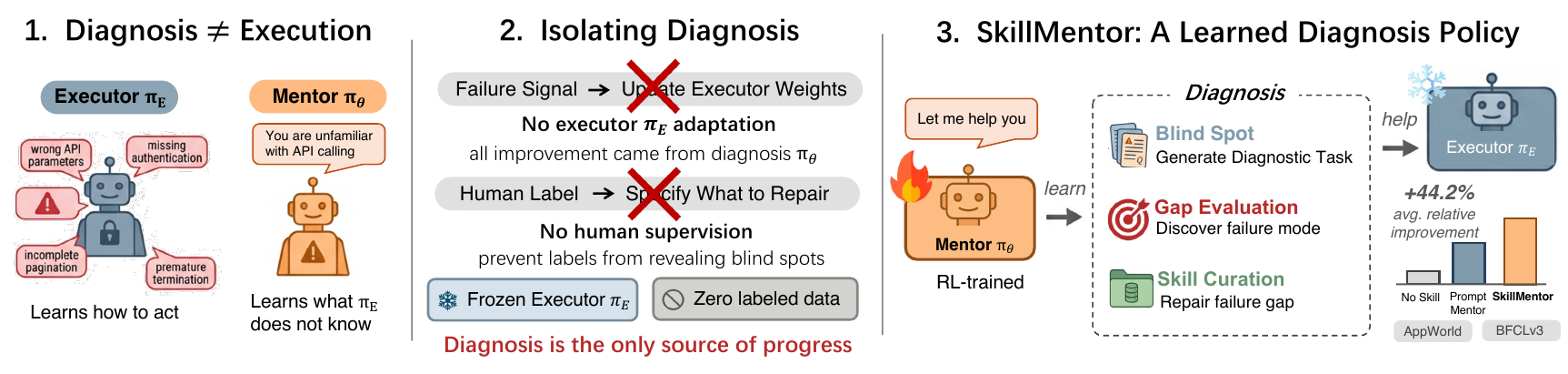}
\caption{ Diagnosis and execution are distinct capabilities. To study diagnosis in isolation, we freeze the executor and remove human supervision, making diagnosis the only source of improvement. SkillMentor trains a Mentor policy to discover blind spots and build an external skill repository, enabling self-evolution without updating executor weights.}
\label{fig:main}
\end{figure}
\section{Introduction}
Unlike an executor, which learns how to act, a mentor learns how to diagnose. Agent self-evolution has focused almost exclusively on the former, overlooking an equally important prerequisite: discovering an agent's blind spots, the recurrent failures that limit its capabilities. This capability, diagnosis, is fundamentally distinct from execution.

Current methods largely bypass diagnosis. Methods that train executor parameters, such as SkillRL~\citep{xia2026skillrl}, SKILL0~\citep{lu2026skill0}, and AgentEvolver~\citep{zhai2025agentevolver}, focus on learning how to act rather than learning to diagnose. Methods that manage external skill libraries, such as SkillOS~\citep{ouyang2026skillos} and SkillOPT~\citep{yang2026skillopt}, rely on pre-defined datasets or validation splits to determine what to repair. Test-time memory frameworks such as ReasoningBank~\citep{ouyang2025reasoningbank} and Reflexion~\citep{shinn2023reflexion} extract skills by prompting a strong model at inference time, but the strong model itself remains static and does not improve through experience. In all cases, diagnosis remains an engineered procedure rather than a capability explicitly optimized through learning.

We therefore ask: can an agent learn to diagnose another agent's blind spots and repair them? Since diagnosis is conceptually distinct from execution, we study a setting that explicitly separates the two: the executor is frozen and no human-labeled data is available.
These constraints ensure that adaptation occurs only through the Mentor and the external skill repository: 
1) If the executor were trainable, performance gains could arise from either the  diagnosis or the executor's parameter updates, making repairs no longer uniquely attributable to diagnosis. This introduces an additional adaptation pathway that prevents diagnosis from being isolated as the capability under study. We therefore externalize adaptation through an explicit skill repository, where each curated skill makes the diagnosed blind spot observable and inspectable rather than implicitly absorbed into the executor's parameters.
2) If labeled data were available, supervision would explicitly specify what to repair, bypassing blind-spot discovery altogether. By removing both shortcuts, the system is forced to identify and repair recurrent failures through diagnosis, allowing it to be studied  independently.

Under these constraints, we formalize the target of diagnosis as a \textbf{blind spot}: a recurrent failure mode that consistently degrades performance on a subset of tasks yet can be mitigated by an external procedural skill. Examples include incorrect API parameter handling, missing authentication and premature termination.
Blind spots are inherently dynamic. As corrective skills are added, the executor's failure distribution shifts and new blind spots emerge. Blind-spot discovery and curation therefore form a positive feedback loop: discovering blind spots leads to better skills, while better skills expose previously hidden blind spots. Optimizing either process in isolation misses their interdependence.

Based on this insight, we propose \textbf{SkillMentor}, which trains a Mentor policy $\pi_\theta$ via reinforcement learning to actively discover an executor's blind spots and repair them. SkillMentor operates in three stages: (1) \textbf{Blind-Spot Discovery}, where the Mentor generates diagnostic tasks through environment interaction; (2) \textbf{Gap Evaluation}, where the Mentor identifies executor's failure modes by measuring diagnostic gaps; and (3) \textbf{Skill Curation}, where the Mentor summarizes corrective skills and retains them only if they improve executor performance.
By jointly optimizing discovery and curation, SkillMentor forms a self-evolving feedback loop: better discovery yields better skills, while better skills reveal previously hidden blind spots. Rather than treating diagnosis as a fixed procedure, SkillMentor learns it as an adaptive policy through experience.

We evaluate SkillMentor on AppWorld and BFCLv3 under frozen-executor and zero-human-data conditions. Across three executors, SkillMentor achieves a 44.2\% average relative improvement over the No Skill baseline and consistently outperforms strong prompt-based mentors such as DeepSeek-V4-Flash. Ablations further reveal a positive feedback loop between discovery and curation: discovering blind spots improves skill quality, while better skills expose new blind spots. Together, these results suggest that diagnosis itself can be learned, enabling agent self-evolution without updating executor weights or relying on human-curated data.
\section{Related Work}

Our work intersects three research themes: learning to execute, optimizing external skills, and learning without supervision. Recent surveys~\citep{fang2025survey} and benchmarks~\citep{wu2024streambench,he2026memoryarena} highlight growing interest in these directions.

\noindent\textbf{Learning to Execute}:
A dominant line of work improves agent performance by updating executor parameters. SkillRL~\citep{xia2026skillrl} trains a Qwen2.5-7B executor via GRPO to use a skill repository distilled by a strong teacher. SKILL0~\citep{lu2026skill0} further internalizes these skills into the executor's parameters through progressive skill withdrawal. AgentEvolver~\citep{zhai2025agentevolver} autonomously generates tasks through self-questioning and trains the executor via self-play, while SAGE~\citep{wang2025sage} exploits sequential rollouts so that earlier skills benefit later tasks in the same chain. D2Skill~\citep{tu2026d2skill} and ARISE~\citep{li2026arise} further augment executor training with evolving skill banks.
Although these methods differ in how they acquire or organize skills, they all aim to improve the executor rather than learn how to diagnose it. Any analysis of failures is performed only to produce better training signals for executor optimization, not to learn a diagnosis policy. As a result, they never treat identifying the executor's blind spots as the learning objective.

\noindent\textbf{External Skill Optimization}:
A second line of work keeps the executor frozen and instead improves the external knowledge it consumes. SkillOS~\citep{ouyang2026skillos} trains an RL-based Curator to manage an evolving SkillRepo through insert, update, and delete operations on grouped task streams. SkillOPT~\citep{yang2026skillopt} formulates skill editing as a controllable text-space optimization process using bounded edits and held-out validation gates. Other methods explore verification-driven curation~\citep{zhang2026coevoskills}, multi-agent elite pools~\citep{alzubi2026evoskill}, dual-loop skill injection~\citep{yang2026autoskill}, memory-specific skill optimization~\citep{zhang2026memskill}, and automated tool creation for agents~\citep{qiu2025alita,liang2026skillnet}. While these methods improve external skills, they assume that diagnostic signals are already available, such as dataset splitting, validation procedures, or verification mechanisms. Consequently, they optimize how skills are refined after failures are identified, rather than learning to autonomously discover what the executor does not know under zero human supervision.

\noindent\textbf{Learning without Supervision}:
Recent work has explored eliminating human-provided data entirely. Agent0~\citep{xia2025agent0}, Tool-R0~\citep{acikgoz2026toolr0}, Absolute Zero~\citep{zhao2025absolutezero}, and EvoEnv~\citep{shi2026evoenv} autonomously generate training tasks, but use these tasks to optimize executor parameters rather than learning a policy that identifies and repairs the executor's blind spots. ReasoningBank~\citep{ouyang2025reasoningbank} and Reflexion~\citep{shinn2023reflexion} instead keep the executor frozen and require no training data, but rely on a static stronger model to extract or refine skills through inference-time prompting, leaving the diagnosis process itself fixed rather than learned. Expel~\citep{zhao2024expel} extracts reusable knowledge from cross-task experience without weight updates, and MemRL~\citep{zhang2026memrl} optimizes episodic memory through runtime reinforcement learning, yet neither trains a dedicated diagnosis policy for autonomously discovering recurrent failure patterns. Consequently, existing unsupervised approaches improve execution, memory, or skill utilization, but diagnosis itself remains an unlearned capability.


\noindent\textbf{Positioning SkillMentor}:
Existing self-evolving agents improve execution by updating the executor, refining external knowledge, or leveraging fixed diagnostic signals, but they do not learn diagnosis itself. In each line, diagnosis is hand-coded, prompted from a strong model, or implicitly absorbed into executor weights, never treated as an optimization target. SkillMentor instead treats diagnosis itself as a learning objective. Rather than asking how an agent should act better, SkillMentor asks how another agent can learn what the executor does not know.
By jointly optimizing blind-spot discovery and skill curation through reinforcement learning, a small Mentor acquires the ability to discover an executor's deficiencies from scratch, turning diagnosis from a fixed pipeline step into an adaptive capability. Since diagnosis is learned rather than prompted, the strong model serves only as an LLM judge during training and is removed at deployment, yielding the lightest strong-model dependency among existing methods as shown in Table~\ref{tab:strong_model}. 

\section{Method}

SkillMentor trains a Mentor policy to diagnose another agent's blind spots. We study this capability under the cleanest setting: the diagnosed agent (Executor) is frozen and no human-labeled data is available. Section~\ref{sec:problem} formalizes blind-spot discovery as a sequential decision problem. Sections~\ref{sec:discovery} to~\ref{sec:curation} detail how the Mentor generates diagnostic tasks, evaluates their diagnostic value, and curates corrective skills. Section~\ref{sec:training} describes the training objective. All prompt templates used by the Mentor are provided in Appendix~\ref{app:prompts}.

\subsection{Problem Formulation}
\label{sec:problem}
\begin{figure}[tp]
\centering
\includegraphics[width=\linewidth]{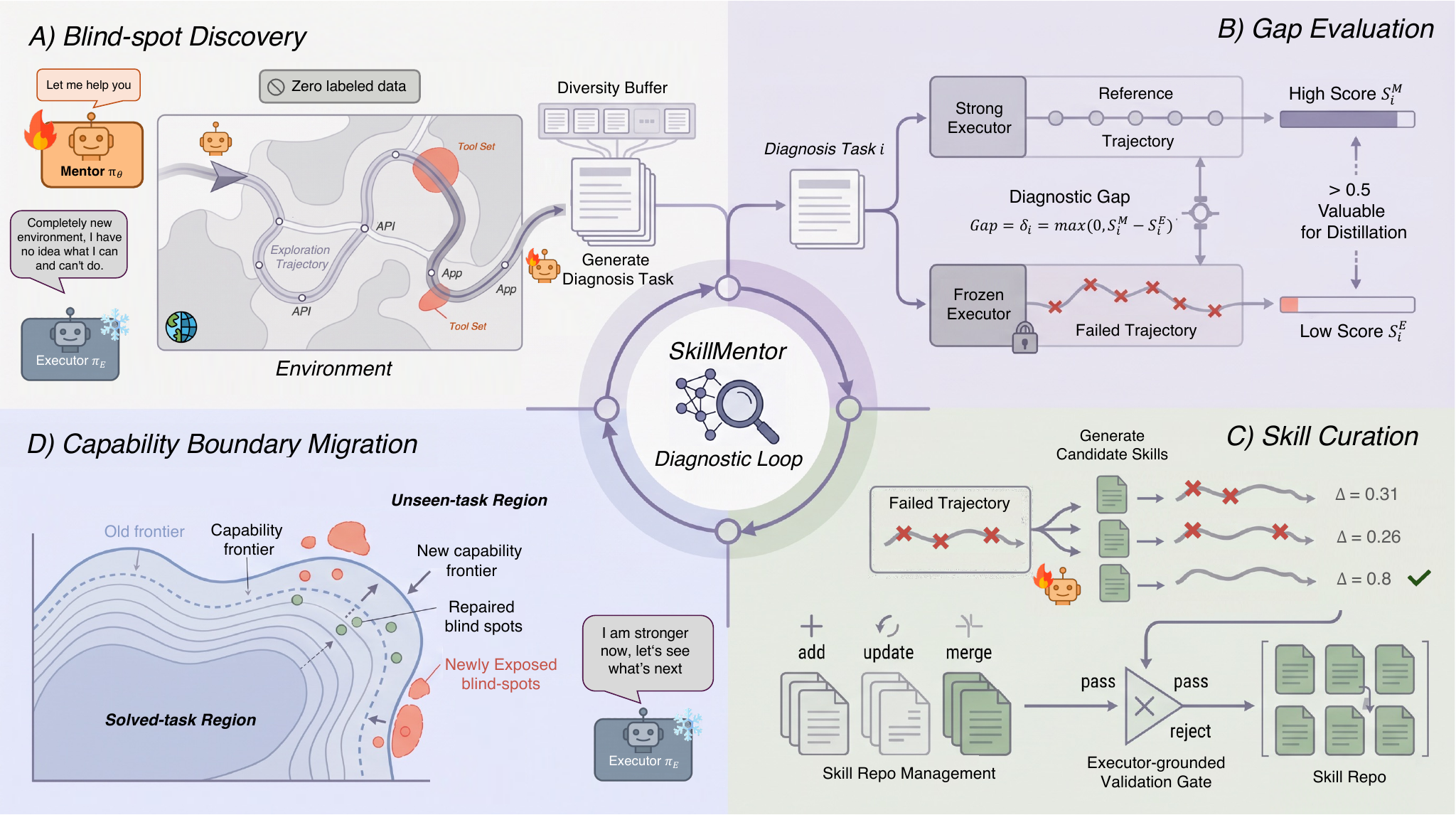}
\caption{ Overall design of SkillMentor.}
\label{fig:verall}
\end{figure}
Unlike prior work that engineers diagnosis through fixed procedures, we formulate diagnosis itself as a policy optimization problem. A blind spot is defined as a tuple $b=(\mathcal{T},c)$, where $\mathcal{T}$ denotes a task on which the Executor $\pi_E$ consistently underperforms, and $c$ is a corrective skill that mitigates the failure.

At training step $t$, the Mentor $\pi_\theta$ observes the environment $\mathcal{E}$ together with the current skill repository $\mathcal{R}_t$, and jointly decides (i) which tasks to diagnose and (ii) how to repair the most informative blind spot. We represent this decision as an action $a_t=(q_t,c_t)$, where $q_t$ is diagnostic task and $c_t$ is a curation decision that converts the diagnosed failure into a corrective skill. If the resulting skill passes validation, it is added to the repository, yielding an updated repository $\mathcal{R}_{t+1}$.

The Mentor receives two rewards: a discovery reward $r_{\text{disc}}$ that measures how informative the generated tasks are, and a curation reward $r_{\text{curate}}$ that measures the utility of the resulting skill. The overall objective is to maximize their cumulative sum:
 \begin{equation}
\max_\theta
\;
\mathbb{E}_{\pi_\theta}
[
\sum_t
\big(
r_{\text{disc}}(q_t)
+
r_{\text{curate}}(c_t)
\big)
]
\end{equation}
Crucially, the two rewards are coupled through the evolving skill repository. Discovery determines which blind spots are exposed, while curation changes which blind spots remain. This dependency creates a discovery-curation feedback loop that motivates joint optimization throughout training.

\subsection{Blind-Spot Discovery}
\label{sec:discovery}

We model the target environment as an interaction sandbox $\mathcal{E} = (\mathcal{X}, \mathcal{A}, \mathcal{P})$, where $\mathcal{X}$ denotes the state space, $\mathcal{A}$ is the set of available tools and APIs exposed by the environment, and $\mathcal{P}(x_{t+1} \mid x_t, a_t)$ defines the transition dynamics. The environment provides no predefined objectives or rewards.

As shown in Figure~\ref{fig:verall} A), at each training step $t$, the Mentor explores $\mathcal{E}$ from an initial state $x_0$ and interacts with the environment for $T$ steps:
  \begin{equation}
  a_t \sim \pi_{\theta}(\cdot \mid x_t, a_{t-1}, x_{t-1}, \dots, a_0, x_0; \mathcal{R}_t), \quad x_{t+1} \sim \mathcal{P}(\cdot \mid x_t, a_t),
  \end{equation}
yielding a single exploration trajectory $\tau = (x_0, a_0, x_1, a_1, \dots, x_T)$. To prevent the exploration from collapsing to a narrow set of behavior patterns, the Mentor maintains a diversity buffer $\mathcal{B}_t$ of recent trajectories and penalizes actions that lead to states already well-covered by the buffer.

From this trajectory, the Mentor synthesizes a batch of $G$ candidate diagnostic tasks $\mathcal{Q}_t = \{q_1, \dots, q_G\}$. Each task $q$ encodes a challenge derived from the observed  environment interaction. All $G$ tasks receive a discovery reward and contribute to GRPO training; the task with the largest diagnostic gap proceeds to Skill Curation, which will be introduced in the next two subsections.




\subsection{Gap Evaluation}
\label{sec:diagnosing}

Each candidate task undergoes dual evaluation. As shown in Figure~\ref{fig:verall} B), given a task $q_i$, a strong reference model $\mathcal{M}$ generates a trajectory $\tau_i^{\text{M}}$ and scores it with $s_i^{\text{M}}\in[0,1]$. The frozen Executor $\pi_E$ attempts the same task with skills retrieved from the current repository $\mathcal{R}_t$, producing trajectory $\tau_i^{\text{E}}$ and score $s_i^{\text{E}}\in[0,1]$ under the same criteria.
The diagnostic gap$\delta_i$ and discovery reward $r_{\text{disc}}$ of $q_i$ is defined as:
\begin{equation}
 r_{\text{disc}}=\delta_i=\max\left(0,
s_i^{\text{M}}-s_i^{\text{E}}
\right)
\end{equation}
The strong model acts as a reference policy rather than  supervision, since only relative performance differences are used and no labels are provided.
The task $q_{max}$ achieving the largest gap $\delta_{\max}$ is selected for Skill Curation. If $\delta_{\max}$ falls below a threshold, Skill Curation is skipped and only Blind-Spot Discovery receives a learning signal.

Because successful skill accumulation naturally shrinks diagnostic gaps over time, a fixed threshold would eventually exclude all tasks from curation. We therefore employ a linearly decaying threshold that starts at $0.5$ and decreases by $0.002$ per step to a floor of $0.2$, tracking the declining gap distribution. We compare this schedule against static and rising alternatives in Section~\ref{sec:evolution}.

\subsection{Skill Curation}
\label{sec:curation}

Skill Curation transforms diagnosed blind spots into reusable corrective skills. As shown in Figure~\ref{fig:verall} C), the Mentor $\pi_\theta$ receives the diagnosed task $q_{max}$, the Executor's failed trajectory $\tau^{\text{E}}$ and relevant retrieved skills from the existing repository.

The Mentor $\pi_\theta$ generates $G$ candidate skills via three operations: \texttt{ADD}, \texttt{UPDATE}, and \texttt{MERGE}, which respectively expand, refine, and compress the repository. All $G$ skills receive a curation reward $r_{\text{curate}}$ and contribute to GRPO training; only the best candidate that passes both validation checks enters the repository.
Each candidate is evaluated along two dimensions.
1) \textbf{Syntactic validity:} The skill must conform to a predefined schema and produce a valid repository operation.
2) \textbf{Executor-grounded utility:} The Executor re-attempts the task with the candidate skill available. Let $c$ denote a candidate skill. The resulting score $s_c$ directly measures whether the skill improves the Executor's behavior.

The improvement over the previous baseline is $\Delta=s_c-s_{\max}^{\text{E}}$,
where $s_{\max}^{\text{E}}$ is the Executor's score recorded during Gap Evaluation. A candidate enters the repository only if it is syntactically valid and achieves a $\Delta > 0.5$.
Then the Skill Curation reward is defined as:
\begin{equation}
r_{\text{curate}}
=
0.3\,f(c)
+
0.7\,\Delta
\end{equation}
Here, $f(c) \in \{0, 1\}$ is a binary format-compliance score: it equals 1 only if the candidate specifies a valid operation type (\texttt{ADD}, \texttt{UPDATE}, or \texttt{MERGE}) and fits a predefined skill format, otherwise $f(c) = 0$. 

To prevent the repository from accumulating obsolete or ineffective skills, we employ a lightweight \texttt{DELETE}  mechanism. For each skill, we track how many times it has been retrieved since insertion and, among those retrievals, how many led to a successful task completion. Any skill whose success rate falls below 0.05 is evicted from the repository. Hyperparameter settings and sensitivity are reported in Appendix~\ref{app:config} and~\ref{app:sensitivity}.

\subsection{Capability Boundary Migration}\label{sec:training}

\textbf{Joint Optimization.}
The Mentor $\pi_\theta$ is trained with GRPO~\citep{shao2024deepseekmath} to jointly optimize two rewards: the discovery reward $r_{\text{disc}}$, which encourages discovering informative blind spots, and the curation reward $r_{\text{curate}}$, which encourages generating useful corrective skills. At each step, $G=8$ tasks and skills are sampled and advantages are normalized within the group. Following the standard GRPO, we maximize the objective
\begin{equation}
J_{\text{GRPO}}(\theta) = J_{\text{disc}} + J_{\text{curate}} - \beta 
       D_{\mathrm{KL}}(\pi_\theta \| \pi_{\mathrm{ref}})
\end{equation}
where $J_{\text{disc}}$ and $J_{\text{curate}}$ are the clipped-surrogate GRPO objectives computed from $r_{\text{disc}}$ and $r_{\text{curate}}$ respectively, each with clipping range $\varepsilon=0.2$. The term $J_{\text{curate}}$ is disabled whenever the diagnostic signal $\delta_{\max}$ is insufficient. Joint optimization is essential because discovery and curation form a positive feedback loop: discovering blind spots produces better skills, while better skills shift the Executor's capability boundary and expose previously hidden blind spots. Training either capability in isolation breaks this loop, we will examine this joint optimization in Section~\ref{sec:ablation}.

\noindent\textbf{Inference and Capability Boundary Migration.}
As shown in Figure~\ref{fig:verall} D), at inference the Executor selects relevant skills from the repository by matching skill descriptions to the current task, and prepends them to its prompt. As skills accumulate, the Executor's capability boundary progressively migrates, tracked by the decaying diagnostic threshold. The repository thus serves as both the product of diagnosis and the engine of self-evolution.

\begin{table}[!t]
\centering
\caption{Main results on AppWorld and BFCLv3. All results are under zero-data, frozen-executor conditions. Mentor variants share the same executor; memory baselines accumulate knowledge across episodes at inference time.}
\label{tab:main}
\setlength{\tabcolsep}{1.6mm}{
\scalebox{0.83}{
\renewcommand{\multirowsetup}{\centering}
\begin{tabular}{lcccccccc}
\toprule
\multirow{2}{*}{Method} & \multirow{2}{*}{Mentor} & \multicolumn{2}{c}{AppWorld} & \multicolumn{5}{c}{BFCLv3} \\
\cmidrule(lr){3-4} \cmidrule(lr){5-9}
& & Acc & Step & Agentic & M-Turn & S-Turn & Hallu. & Avg Acc \\
\midrule
 \rowcolor{gray!20} \multicolumn{9}{c}{\textit{ Executor $\pi_E$  Qwen3.5-9B}} \\
No Skill  &  --  & 0.300\std{±.010} & 21.0\std{±.009} & 0.452\std{±.018} & 0.488\std{±.016} & 0.515\std{±.015} & 0.458\std{±.017} & 0.478\std{±.015} \\
Reflexion & {\includegraphics[height=2.2ex]{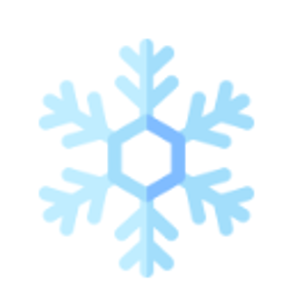}} Qwen3.5-4B & 0.312\std{±.014} & 20.8\std{±.012} & 0.476\std{±.019} & 0.512\std{±.017} & 0.540\std{±.016} & 0.482\std{±.018} & 0.502\std{±.017} \\
MemP & {\includegraphics[height=2.2ex]{image/ice.png}} Qwen3.5-4B & 0.325\std{±.013} & 20.6\std{±.011} & 0.500\std{±.018} & 0.535\std{±.016} & 0.565\std{±.015} & 0.508\std{±.017} & 0.527\std{±.016} \\
ReasoningBank & {\includegraphics[height=2.2ex]{image/ice.png}} Qwen3.5-4B & 0.335\std{±.011} & 20.6\std{±.010} & 0.518\std{±.016} & 0.552\std{±.015} & 0.585\std{±.014} & 0.526\std{±.015} & 0.545\std{±.014} \\
\multirow{4}{*}{SkillMentor} &  \raisebox{-0.5ex}{\includegraphics[height=2.2ex]{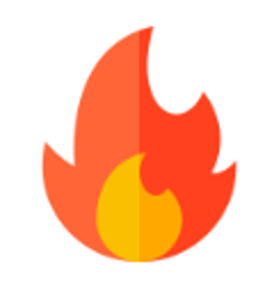}} Qwen3.5-4B   & \textbf{0.410}\std{±.008} & \textbf{18.5}\std{±.007} & \textbf{0.660}\std{±.014} & \textbf{0.695}\std{±.013} & \textbf{0.720}\std{±.012} & \textbf{0.653}\std{±.014} & \textbf{0.682}\std{±.012} \\
 & {\includegraphics[height=2.2ex]{image/ice.png}} Qwen3.5-4B  & 0.340\std{±.010} & 20.5\std{±.009} & 0.528\std{±.016} & 0.562\std{±.015} & 0.592\std{±.014} & 0.534\std{±.015} & 0.554\std{±.014} \\
& {\includegraphics[height=2.2ex]{image/ice.png}} Qwen3.6-Flash  & 0.361\std{±.009} & 19.8\std{±.008} & 0.572\std{±.015} & 0.605\std{±.014} & 0.638\std{±.013} & 0.578\std{±.014} & 0.598\std{±.013} \\
& {\includegraphics[height=2.2ex]{image/ice.png}} DeepSeek-V4-Flash  & 0.379\std{±.008} & 19.2\std{±.008} & 0.610\std{±.014} & 0.642\std{±.013} & 0.675\std{±.012} & 0.613\std{±.014} & 0.635\std{±.012} \\
\midrule
 \rowcolor{gray!20} \multicolumn{9}{c}{\textit{ Executor $\pi_E$  Qwen3.5-4B}} \\
No Skill  &  --  & 0.244\std{±.011} & 24.1\std{±.010} & 0.378\std{±.018} & 0.408\std{±.017} & 0.435\std{±.016} & 0.387\std{±.018} & 0.402\std{±.016} \\
Reflexion & {\includegraphics[height=2.2ex]{image/ice.png}} Qwen3.5-4B & 0.260\std{±.015} & 24.1\std{±.013} & 0.400\std{±.020} & 0.432\std{±.018} & 0.460\std{±.017} & 0.410\std{±.019} & 0.425\std{±.018} \\
MemP & {\includegraphics[height=2.2ex]{image/ice.png}} Qwen3.5-4B & 0.275\std{±.014} & 24.0\std{±.012} & 0.422\std{±.019} & 0.455\std{±.017} & 0.485\std{±.016} & 0.432\std{±.018} & 0.448\std{±.017} \\
ReasoningBank & {\includegraphics[height=2.2ex]{image/ice.png}} Qwen3.5-4B & 0.290\std{±.012} & 24.0\std{±.011} & 0.438\std{±.017} & 0.472\std{±.016} & 0.502\std{±.015} & 0.450\std{±.016} & 0.465\std{±.015} \\
\multirow{4}{*}{SkillMentor} &  \raisebox{-0.5ex}{\includegraphics[height=2.2ex]{image/fire.png}} Qwen3.5-4B   & \textbf{0.351}\std{±.009} & \textbf{22.7}\std{±.008} & \textbf{0.572}\std{±.015} & \textbf{0.605}\std{±.014} & \textbf{0.635}\std{±.013} & \textbf{0.572}\std{±.015} & \textbf{0.596}\std{±.013} \\
 & {\includegraphics[height=2.2ex]{image/ice.png}} Qwen3.5-4B  & 0.295\std{±.011} & 24.0\std{±.010} & 0.448\std{±.017} & 0.482\std{±.016} & 0.510\std{±.015} & 0.460\std{±.016} & 0.475\std{±.015} \\
& {\includegraphics[height=2.2ex]{image/ice.png}} Qwen3.6-Flash  & 0.318\std{±.010} & 23.4\std{±.009} & 0.492\std{±.016} & 0.528\std{±.015} & 0.558\std{±.014} & 0.502\std{±.015} & 0.520\std{±.014} \\
& {\includegraphics[height=2.2ex]{image/ice.png}} DeepSeek-V4-Flash  & 0.335\std{±.009} & 22.9\std{±.009} & 0.525\std{±.015} & 0.560\std{±.014} & 0.590\std{±.013} & 0.533\std{±.015} & 0.552\std{±.013} \\
\midrule
 \rowcolor{gray!20} \multicolumn{9}{c}{\textit{ Executor $\pi_E$  DeepSeek-R1-Distill-Qwen-7B}} \\
No Skill  &  --  & 0.208\std{±.012} & 27.5\std{±.011} & 0.332\std{±.020} & 0.365\std{±.018} & 0.385\std{±.017} & 0.338\std{±.019} & 0.355\std{±.018} \\
Reflexion & {\includegraphics[height=2.2ex]{image/ice.png}} Qwen3.5-4B & 0.224\std{±.016} & 27.2\std{±.014} & 0.352\std{±.022} & 0.385\std{±.019} & 0.408\std{±.018} & 0.358\std{±.020} & 0.375\std{±.019} \\
MemP & {\includegraphics[height=2.2ex]{image/ice.png}} Qwen3.5-4B & 0.238\std{±.015} & 26.8\std{±.013} & 0.372\std{±.021} & 0.405\std{±.018} & 0.430\std{±.017} & 0.380\std{±.019} & 0.396\std{±.018} \\
ReasoningBank & {\includegraphics[height=2.2ex]{image/ice.png}} Qwen3.5-4B & 0.251\std{±.013} & 26.6\std{±.012} & 0.388\std{±.019} & 0.422\std{±.017} & 0.446\std{±.016} & 0.396\std{±.017} & 0.413\std{±.016} \\ 
\multirow{4}{*}{SkillMentor} &  \raisebox{-0.5ex}{\includegraphics[height=2.2ex]{image/fire.png}} Qwen3.5-4B   & \textbf{0.305}\std{±.010} & \textbf{24.0}\std{±.009} & \textbf{0.498}\std{±.016} & \textbf{0.530}\std{±.015} & \textbf{0.562}\std{±.014} & \textbf{0.502}\std{±.016} & \textbf{0.523}\std{±.014} \\
 & {\includegraphics[height=2.2ex]{image/ice.png}} Qwen3.5-4B  & 0.255\std{±.012} & 26.5\std{±.011} & 0.395\std{±.018} & 0.428\std{±.017} & 0.452\std{±.016} & 0.405\std{±.018} & 0.420\std{±.016} \\
& {\includegraphics[height=2.2ex]{image/ice.png}} Qwen3.6-Flash  & 0.274\std{±.011} & 25.8\std{±.010} & 0.435\std{±.017} & 0.470\std{±.016} & 0.498\std{±.015} & 0.445\std{±.017} & 0.462\std{±.015} \\
& {\includegraphics[height=2.2ex]{image/ice.png}} DeepSeek-V4-Flash  & 0.290\std{±.010} & 25.2\std{±.010} & 0.468\std{±.016} & 0.502\std{±.015} & 0.532\std{±.014} & 0.478\std{±.016} & 0.495\std{±.014} \\

\bottomrule
\end{tabular}
}
}
\end{table}

\section{Experiments}
\label{sec:experiments}

\subsection{Main Results: Does Learning Improve Diagnosis?}
\label{sec:main}

We evaluate SkillMentor under frozen-executor and zero-human-data constraints to answer a central question: \emph{Does diagnosis benefit from learning, rather than being executed through a fixed prompting procedure?}

To isolate this effect, we compare SkillMentor against prompt-based mentors that share the same architecture, skill format, and retrieval mechanism but receive no RL training. Specifically, we consider a \textbf{No Skill} executor and three frozen mentors: Qwen3.5-4B, Qwen3.6-Flash, and DeepSeek-V4-Flash. This controlled comparison holds the skill infrastructure fixed and isolates the contribution of learning diagnosis itself. We evaluate on two benchmarks: AppWorld~\citep{trivedi-etal-2024-appworld}  for long-horizon planning and BFCLv3~\citep{pmlr-v267-patil25a} for precise function calling.

We select baselines that must share our supervision constraint:
Reflexion~\citep{shinn2023reflexion}, MemP~\citep{fang2025memp}, and ReasoningBank~\citep{ouyang2025reasoningbank} keep the executor frozen and use no labeled data, extracting corrective knowledge purely through inference-time prompting; comparing against them isolates whether learned diagnosis outperforms fixed prompting. SkillOS~\citep{ouyang2026skillos} and SkillOPT~\citep{yang2026skillopt} are excluded rather than overlooked, since their core mechanisms require supervision our constraint disallows.


Table~\ref{tab:main} provides a consistent answer across all three executors: \textbf{learning improves diagnosis}. The RL-trained Mentor outperforms every prompt-based mentor in both environments, demonstrating that diagnosis benefits from optimization rather than a fixed prompting procedure.
On AppWorld, SkillMentor reduces interaction steps by 1.4--3.5 per task, while BFCLv3 exhibits comparable increase. Relative to the \textbf{No Skill} baseline, accuracy improves by 44.2\% on average across executors.
Two findings further support this conclusion:

\textbf{Learning beats scale.} The RL-trained 4B Mentor consistently outperforms DeepSeek-V4-Flash, a substantially larger model used only via prompting. This suggests that diagnosis is not merely a by-product of model scale; targeted RL training can produce a smaller Mentor that surpasses significantly larger prompted models.

\textbf{Weaker executors benefit more from diagnosis.} The relative improvement is largest for DeepSeek-R1-Distill-Qwen-7B (+46.6\%) and smallest for Qwen3.5-9B (+36.7\%). This trend is consistent with our formulation of blind spots as procedural deficiencies: weaker executors expose larger capability gaps and therefore offer more opportunities for discovery and curation, whereas stronger executors already possess much of this procedural knowledge.

Overall, these results suggest that diagnosis is a learnable capability rather than a fixed prompting procedure. While stronger prompted models improve performance, an RL-trained Mentor consistently delivers larger gains despite having fewer parameters and no inference-time access to frontier models. Taken together, these findings establish diagnosis as a trainable capability.

\subsection{Ablation: Why Is Joint Optimization Necessary?}
\label{sec:ablation}

\begin{wraptable}{r}{0.5\linewidth}
\vspace{-15pt}
\centering
\caption{Ablation of discovery-curation coupling.}
\label{tab:reward}
\scalebox{0.83}{
\setlength{\tabcolsep}{1mm}{
\begin{tabular}{lccccc}
\toprule
& Discovery & Curation & Shared & Acc & Step \\
\midrule
A (Ours) & \checkmark & \checkmark & \checkmark & \textbf{0.351} & \textbf{22.7} \\
B        & --         & \checkmark & \checkmark & 0.292 & 25.4 \\
C        & \checkmark & --         & \checkmark & 0.303 & 23.8 \\
D        & \checkmark & \checkmark & --         & 0.336 & 23.5 \\
\bottomrule
\end{tabular}}}
\end{wraptable}

SkillMentor is built on a central hypothesis: discovery and curation should be optimized jointly because they form a feedback loop through the evolving skill repository. We check whether this coupling is truly necessary on AppWorld with Qwen3.5-4B executor.

Table~\ref{tab:reward} shows that the answer is affirmative. The full model (A) achieves the best performance (Acc $0.351$, 22.7 steps). Removing the discovery reward (B) drops accuracy to $0.292$, the largest decline among all ablations, while removing the curation reward (C) drops accuracy to $0.303$, both substantially below the full model and above the No Skill baseline ($0.244$, 24.1 steps). Training with separate parameters (D) reaches $0.336$, trailing the shared Mentor by a clear margin. Joint optimization therefore outperforms every decoupled variant, confirming that discovery and curation reinforce each other through shared representations.

Two findings emerge. \textbf{First, discovery provides the primary learning signal}: without it, the Mentor cannot target informative failures, degrading even the curation signal. \textbf{Second, the two objectives share representations}: identifying where the executor fails and articulating how to fix it draw on overlapping knowledge, so a shared policy outperforms two separate ones.

Overall, these results support our discovery-curation co-evolution hypothesis: jointly optimizing  is more effective than decomposing discovery and curation into independent components.

\subsection{Transferability: Are Learned Skills Executor-Specific?}
\label{sec:transfer}

SkillMentor stores knowledge as an external Markdown repository rather than in executor weights. We therefore ask whether the learned skills are executor-specific artifacts or transferable knowledge.
To answer this question, we train a skill repository using one executor and directly deploy it on another without retraining. Table~\ref{tab:transfer} reports the full $8{\times}8$ transfer matrix. Two patterns emerge:

\textbf{Stronger-to-weaker transfer is limited}: when skills learned from a stronger executor are deployed on a weaker one (green cells), performance is consistently below the weaker executor's self-trained result. Strong executors expose fewer blind spots during training, so their repositories omit procedural patterns that weaker executors still require.
\begin{table}[tp]
\centering
\caption{Cross-executor transfer (Acc). Rows: executor used to train the Mentor. Columns: executor the skills are deployed on. Pink: self-trained (diagonal). Green: skills from a stronger executor transferred to a weaker one. Yellow: skills from a weaker executor transferred to a stronger one. All values exceed the No Skill baseline. DS-R1-7B stands for DeepSeek-R1-Distill-Qwen-7B}
\label{tab:transfer}

\setlength{\tabcolsep}{2.5mm}
\scalebox{0.86}{
\begin{tabular}{ll|cccc|cccc}
\toprule
\multicolumn{2}{c|}{\small AppWorld} & \multicolumn{4}{c|}{Qwen3.5} & \multicolumn{4}{c}{7B-Class} \\
\cmidrule(lr){1-2} \cmidrule(lr){3-6} \cmidrule(lr){7-10}
Train \textbackslash\ Eval & No Skill & 9B & 4B & 2B & 0.6B & MiMo & Gemma & DS-R1-7B & Mistral \\
\midrule
Qwen3.5-9B      & 0.300 & \cellcolor{diagpink}\textbf{0.410} & \cellcolor{transgreen}0.313 & \cellcolor{transgreen}0.243 & \cellcolor{transgreen}0.171 & \cellcolor{transgreen}0.286 & \cellcolor{transgreen}0.241 & \cellcolor{transgreen}0.273 & \cellcolor{transgreen}0.263 \\
Qwen3.5-4B      & 0.244 & \cellcolor{transyellow}0.403 & \cellcolor{diagpink}\textbf{0.351} & \cellcolor{transgreen}0.272 & \cellcolor{transgreen}0.149 & \cellcolor{transyellow}0.335 & \cellcolor{transgreen}0.237 & \cellcolor{transgreen}0.220 & \cellcolor{transgreen}0.218 \\
Qwen3.5-2B      & 0.231 & \cellcolor{transyellow}0.412 & \cellcolor{transyellow}0.352 & \cellcolor{diagpink}\textbf{0.320} & \cellcolor{transgreen}0.166 & \cellcolor{transyellow}0.341 & \cellcolor{transgreen}0.242 & \cellcolor{transgreen}0.219 & \cellcolor{transgreen}0.257 \\
Qwen3.5-0.6B    & 0.136 & \cellcolor{transyellow}0.413 & \cellcolor{transyellow}0.348 & \cellcolor{transyellow}0.314 & \cellcolor{diagpink}\textbf{0.243} & \cellcolor{transyellow}0.347 & \cellcolor{transyellow}0.247 & \cellcolor{transyellow}0.298 & \cellcolor{transyellow}0.269 \\
\midrule
MiMo-7B         & 0.262 & \cellcolor{transyellow}0.416 & \cellcolor{transgreen}0.310 & \cellcolor{transgreen}0.301 & \cellcolor{transgreen}0.213 & \cellcolor{diagpink}\textbf{0.340} & \cellcolor{transgreen}0.237 & \cellcolor{transgreen}0.298 & \cellcolor{transgreen}0.230 \\
Gemma-7B        & 0.218 & \cellcolor{transyellow}0.411 & \cellcolor{transyellow}0.356 & \cellcolor{transyellow}0.322 & \cellcolor{transgreen}0.225 & \cellcolor{transyellow}0.341 & \cellcolor{diagpink}\textbf{0.250} & \cellcolor{transgreen}0.276 & \cellcolor{transgreen}0.208 \\
DS-R1-7B  & 0.208 & \cellcolor{transyellow}0.406 & \cellcolor{transyellow}0.348 & \cellcolor{transyellow}0.313 & \cellcolor{transgreen}0.167 & \cellcolor{transyellow}0.334 & \cellcolor{transyellow}0.246 & \cellcolor{diagpink}\textbf{0.305} & \cellcolor{transgreen}0.246 \\
Mistral-7B      & 0.195 & \cellcolor{transyellow}0.408 & \cellcolor{transyellow}0.349 & \cellcolor{transyellow}0.315 & \cellcolor{transgreen}0.171 & \cellcolor{transyellow}0.347 & \cellcolor{transyellow}0.252 & \cellcolor{transyellow}0.307 & \cellcolor{diagpink}\textbf{0.275} \\

\end{tabular}}
\scalebox{0.86}{
\begin{tabular}{ll|cccc|cccc}
\toprule
\multicolumn{2}{c|}{\small BFCLv3} & \multicolumn{4}{c|}{Qwen3.5} & \multicolumn{4}{c}{7B-Class} \\
\cmidrule(lr){1-2} \cmidrule(lr){3-6} \cmidrule(lr){7-10}
Train \textbackslash\ Eval & No Skill & 9B & 4B & 2B & 0.6B & MiMo & Gemma & DS-R1-7B & Mistral \\
\midrule
Qwen3.5-9B      & 0.478 & \cellcolor{diagpink}\textbf{0.682} & \cellcolor{transgreen}0.483 & \cellcolor{transgreen}0.395 & \cellcolor{transgreen}0.285 & \cellcolor{transgreen}0.423 & \cellcolor{transgreen}0.459 & \cellcolor{transgreen}0.403 & \cellcolor{transgreen}0.378 \\
Qwen3.5-4B      & 0.402 & \cellcolor{transyellow}0.686 & \cellcolor{diagpink}\textbf{0.596} & \cellcolor{transgreen}0.455 & \cellcolor{transgreen}0.393 & \cellcolor{transgreen}0.465 & \cellcolor{transgreen}0.406 & \cellcolor{transgreen}0.457 & \cellcolor{transgreen}0.353 \\
Qwen3.5-2B      & 0.350 & \cellcolor{transyellow}0.690 & \cellcolor{transyellow}0.593 & \cellcolor{diagpink}\textbf{0.545} & \cellcolor{transgreen}0.315 & \cellcolor{transyellow}0.566 & \cellcolor{transgreen}0.364 & \cellcolor{transyellow}0.526 & \cellcolor{transgreen}0.449 \\
Qwen3.5-0.6B    & 0.250 & \cellcolor{transyellow}0.679 & \cellcolor{transyellow}0.605 & \cellcolor{transyellow}0.546 & \cellcolor{diagpink}\textbf{0.410} & \cellcolor{transyellow}0.567 & \cellcolor{transyellow}0.504 & \cellcolor{transyellow}0.521 & \cellcolor{transyellow}0.475 \\
\midrule
MiMo-7B         & 0.370 & \cellcolor{transyellow}0.672 & \cellcolor{transyellow}0.590 & \cellcolor{transgreen}0.535 & \cellcolor{transgreen}0.336 & \cellcolor{diagpink}\textbf{0.562} & \cellcolor{transgreen}0.413 & \cellcolor{transgreen}0.500 & \cellcolor{transgreen}0.412 \\
Gemma-7B        & 0.330 & \cellcolor{transyellow}0.692 & \cellcolor{transyellow}0.593 & \cellcolor{transyellow}0.539 & \cellcolor{transgreen}0.391 & \cellcolor{transyellow}0.555 & \cellcolor{diagpink}\textbf{0.505} & \cellcolor{transyellow}0.529 & \cellcolor{transgreen}0.411 \\
DS-R1-7B  & 0.355 & \cellcolor{transyellow}0.692 & \cellcolor{transyellow}0.605 & \cellcolor{transgreen}0.477 & \cellcolor{transgreen}0.265 & \cellcolor{transyellow}0.552 & \cellcolor{transgreen}0.414 & \cellcolor{diagpink}\textbf{0.523} & \cellcolor{transgreen}0.445 \\
Mistral-7B      & 0.310 & \cellcolor{transyellow}0.674 & \cellcolor{transyellow}0.595 & \cellcolor{transyellow}0.537 & \cellcolor{transgreen}0.304 & \cellcolor{transyellow}0.552 & \cellcolor{transyellow}0.502 & \cellcolor{transyellow}0.523 & \cellcolor{diagpink}\textbf{0.480} \\
\bottomrule
\end{tabular}}

\end{table}

\textbf{Weaker-to-stronger transfer is highly effective}: when skills learned from a weaker executor are transferred to a stronger one (yellow cells), performance often approaches, and occasionally exceeds, the stronger executor's self-trained result. Weaker executors expose more blind spots, producing broader repositories that remain useful to stronger models.

Importantly, every transferred repository outperforms the corresponding No Skill baseline, while diagonal self-trained performance remains competitive. Overall, these results suggest that SkillMentor learns portable diagnostic knowledge: once a blind spot is repaired, its corrective skill can be reused by unseen executors without retraining.

\subsection{Evolution Dynamics: How Do Blind Spots Change Over Time?}
\label{sec:evolution}

Successful skill accumulation changes the Executor itself: tasks that once exposed large failures become easier, causing the blind-spot distribution to shift over time. We therefore ask how diagnosis should adapt throughout training.

\begin{figure}[h]
\centering
\includegraphics[width=0.24\linewidth]{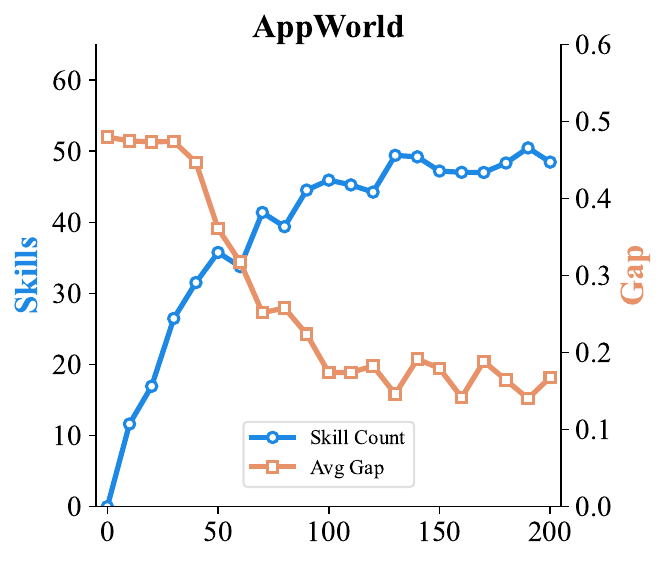}\hspace{0pt}
\includegraphics[width=0.24\linewidth]{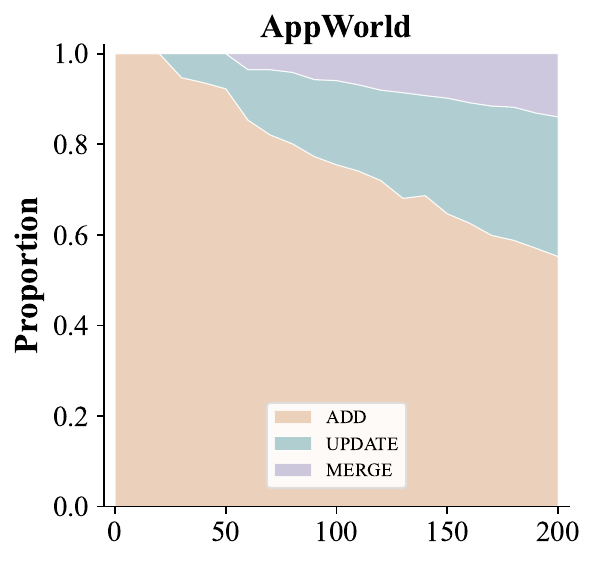}\hspace{0pt}
\includegraphics[width=0.24\linewidth]{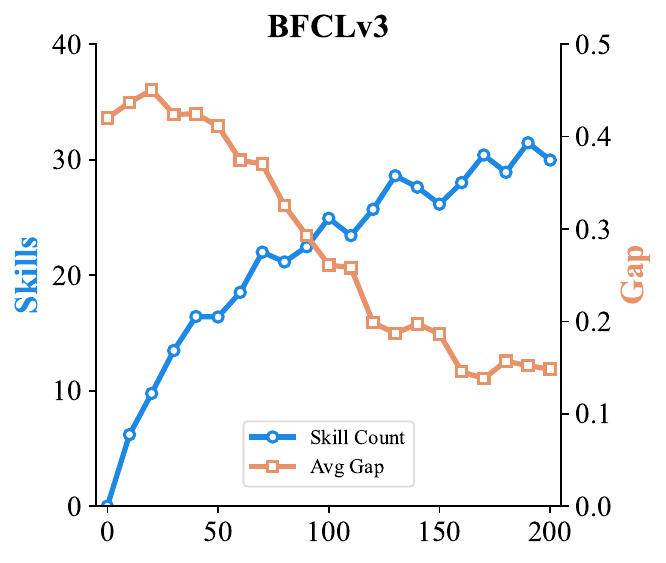}\hspace{0pt}
\includegraphics[width=0.24\linewidth]{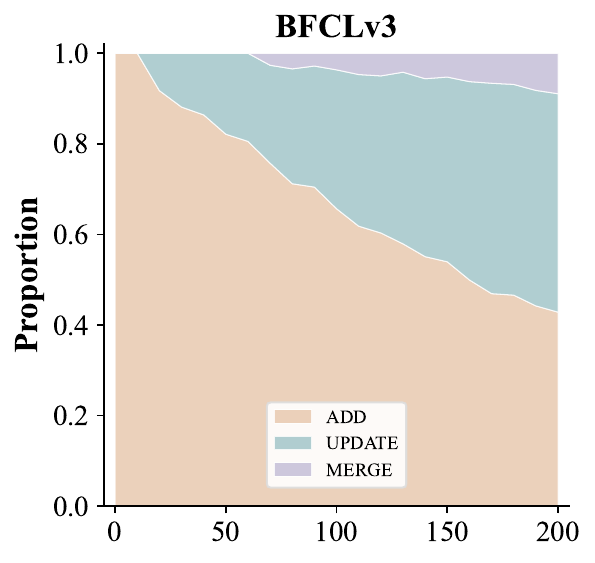}
\caption{Skill repository evolution on AppWorld and BFCLv3 with Qwen3.5-4B. Left two: skill count (blue) and average diagnostic gap (orange), and management action proportions on AppWorld. Right two: the same metrics on BFCLv3.}
\label{fig:library}
\end{figure}

Figure~\ref{fig:library} shows the evolution of the skill repository and the average diagnostic gap. During the first 50 training steps, the repository expands rapidly while the average gap drops from 0.48 to 0.17, indicating that previously severe blind spots are progressively repaired. After step 100, both curves begin to stabilize, suggesting that the Executor has reached a new capability boundary.

\begin{figure}[h]
\centering
\includegraphics[width=0.4\linewidth]{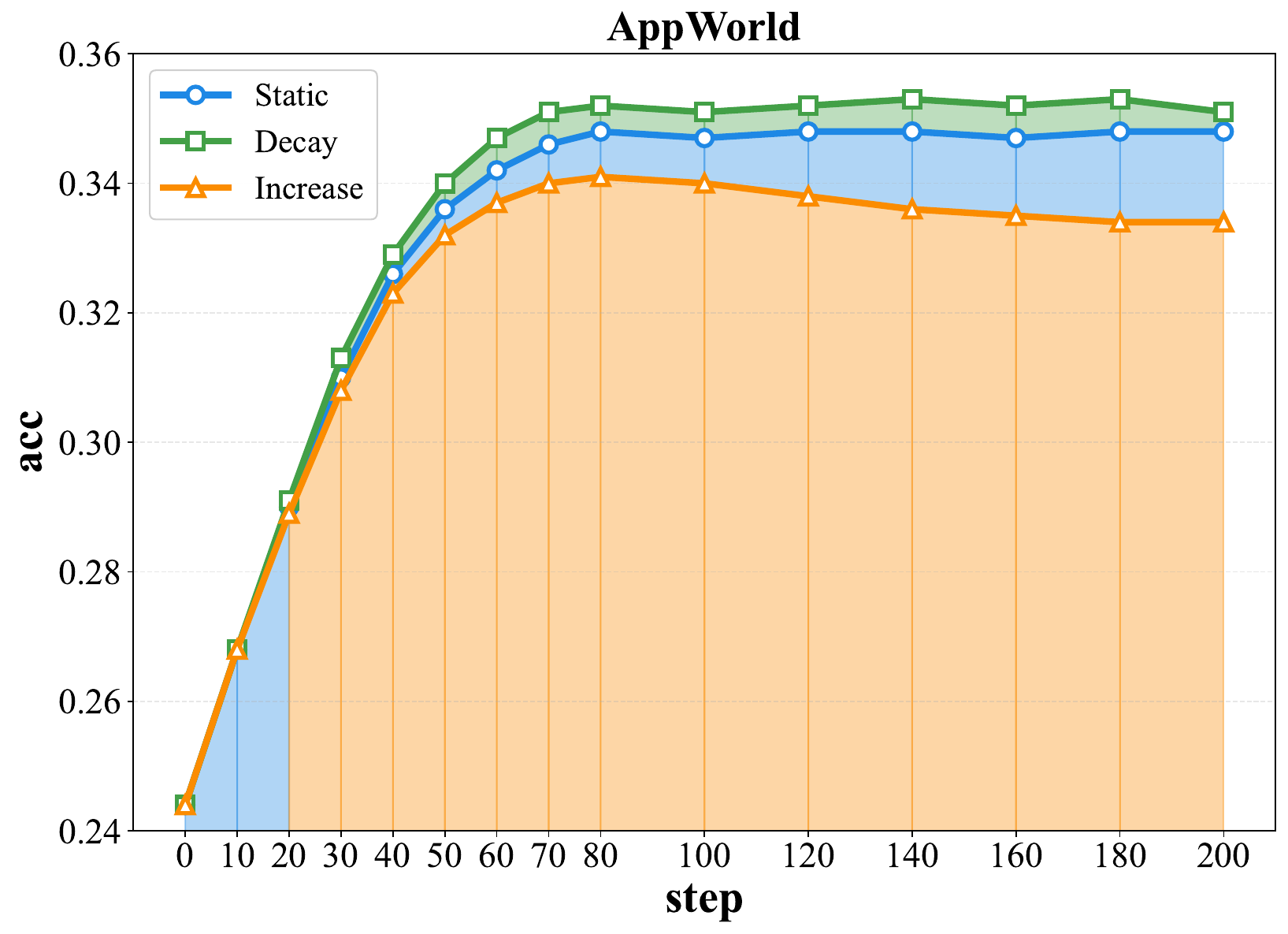}\hspace{12pt}
\includegraphics[width=0.4\linewidth]{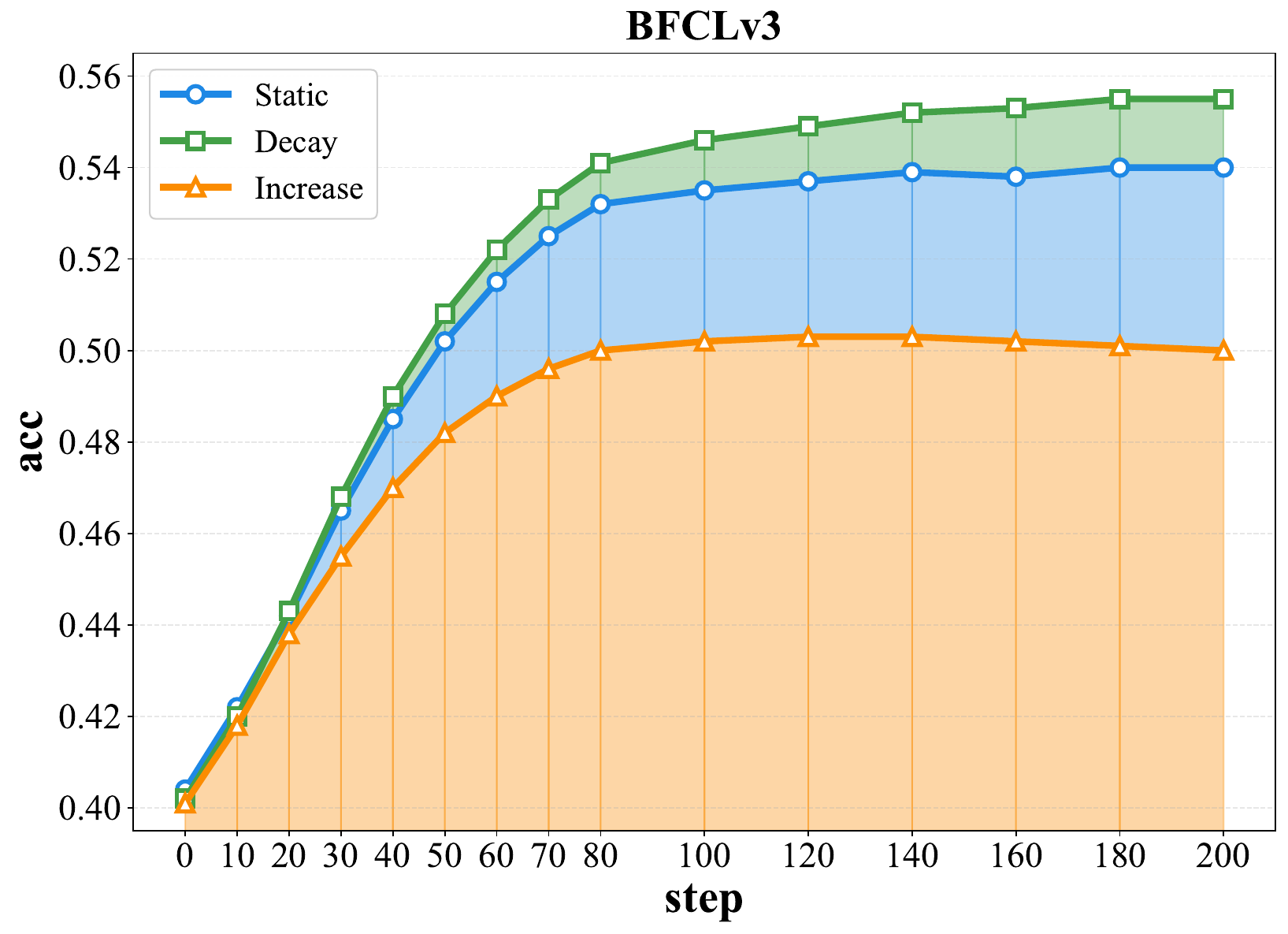}
\caption{Results of three threshold schedules on AppWorld (left) and BFCLv3 (right).}
\label{fig:threshold}
\end{figure}

This shrinking gap distribution directly motivates adaptive diagnosis. We compare three diagnostic threshold schedules on Qwen3.5-4B executor: a static threshold (0.5), a linearly decaying threshold (0.5$\rightarrow$0.2, $\Delta=-0.002$/step), and a linearly rising threshold (0.5$\rightarrow$0.8, $\Delta=+0.002$/step). Figure~\ref{fig:threshold} shows that all schedules behave similarly during early training. After step 100, however, the rising schedule plateaus because progressively fewer tasks satisfy the increasingly strict criterion, while the static schedule eventually saturates. In contrast, the decaying schedule continues to improve by tracking the naturally shrinking gap distribution. The final accuracies are 0.351 (decaying), 0.348 (static), and 0.334 (rising).

Figure~\ref{fig:library} also reveals a shift in curation behavior. Early training is dominated by \texttt{ADD} operations because the repository is initially empty. \texttt{UPDATE} operations increase after step 30, and \texttt{MERGE} operations emerge after step 60, indicating a transition from acquiring new skills to refining and consolidating existing knowledge.

Overall, these results suggest that blind spots are \emph{non-stationary}: repairing existing failures continuously reshapes where future diagnosis should focus, requiring progressive exploration over time.

\subsection{Interpretability: What Does the Mentor Learn?}

\begin{figure}[h]
\centering
\includegraphics[width=\linewidth]{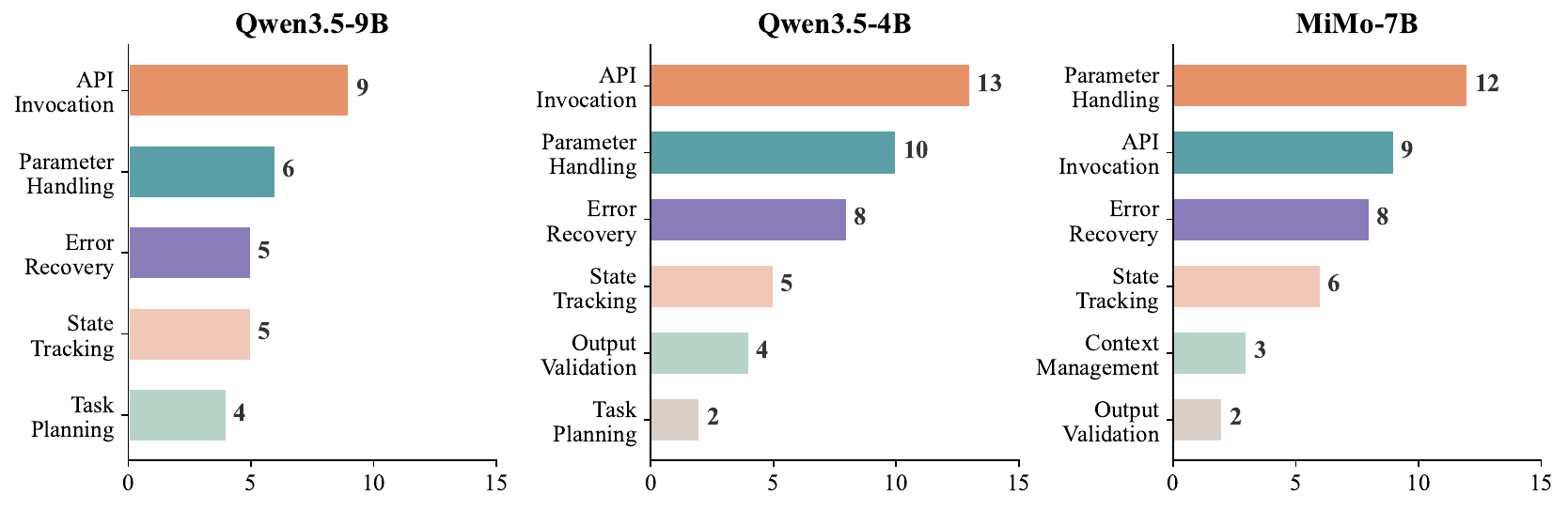}
\caption{Skill comparison across three executors on AppWorld. Left: Qwen3.5-9B (5 categories, 29 skills). Center: Qwen3.5-4B (6 categories, 42 skills). Right: MiMo-7B (6 categories, 40 skills).}
\label{fig:composition}
\end{figure}

Because skills are stored as human-readable Markdown files, the repository itself provides an interpretable view of executor capabilities. We therefore ask: what kinds of blind spots does the Mentor discover?
Figure~\ref{fig:composition} compares the final skill composition of Qwen3.5-9B, Qwen3.5-4B, and MiMo-7B. Two patterns emerge:

\label{sec:interpretability}
\textbf{Within the same model family, smaller executors require more procedural support.} Qwen3.5-4B accumulates substantially more skills than Qwen3.5-9B (42 vs.\ 29) and introduces an additional \textit{output validation} category. The extra skills are concentrated in API invocation and parameter handling, suggesting that precision degrades before high-level reasoning as model capacity decreases.

\textbf{Different model families exhibit distinct blind spots.} MiMo-7B is dominated by parameter-handling deficiencies and uniquely requires a dedicated context-management category that is absent from both Qwen models.

These findings suggest that SkillMentor learns more than a collection of memories: the repository serves as an interpretable map of executor deficiencies. Unlike methods that absorb knowledge into model weights, the discovered blind spots can be directly inspected, edited, and transferred. Representative skills from the final repository are shown in Appendix~\ref{app:skill_examples}, and skill rescue case studies are provided in Appendix~\ref{app:rescued}.

\subsection{Robustness: How Dependent Is SkillMentor on Strong Models?}
\label{sec:judge}
A major concern for self-evolving agents is their reliance on frontier models, which directly affects cost, privacy, and deployment. We therefore ask how sensitive SkillMentor is to the choice of strong model.
Table~\ref{tab:strong_model} first compares strong-model usage across methods. Unlike prior approaches that assign multiple responsibilities to a frontier model (e.g., attribute annotation), SkillMentor restricts its role to a single LLM judge during training and removes it entirely at deployment.

\begin{table}[H]
\centering
\caption{Strong model dependency across agent self-evolution methods. More stars indicate a heavier reliance on the strong model, measured by the number of distinct roles it serves.}
\label{tab:strong_model}
\setlength{\tabcolsep}{1.3mm}
\scalebox{0.82}{
\begin{tabular}{lllc}
\toprule
Method & Strong Model & Roles  & Dependency \\
\midrule
SkillMentor & DeepSeek-V4-Flash & LLM judge  & \textcolor{gold}{\ding{73}} \\
SkillRL & OpenAI o3 & Trajectory Distillation; Skill Proposal & \textcolor{gold}{\ding{73}\ding{73}} \\
SKILL0 & (inherits from SkillRL) & Same as SkillRL;  & \textcolor{gold}{\ding{73}\ding{73}} \\
SkillOS & Gemini-2.5-Pro & Task Attribute Annotation; Content Scoring & \textcolor{gold}{\ding{73}\ding{73}} \\
SkillOpt & GPT-5.5 & Trajectory Analysis; Meta-skill Update & \textcolor{gold}{\ding{73}\ding{73}} \\
AgentEvolver & Qwen-MAX & Step Attribution; Environment Exploration; Summarization & \textcolor{gold}{\ding{73}\ding{73}\ding{73}} \\
\bottomrule
\end{tabular}}
\end{table}

\begin{table}[H]
\centering
\caption{SkillMentor with different LLM judges on AppWorld. Flash-tier judges achieve near-optimal performance at a fraction of the cost (200 steps, Alibaba Cloud, China region).}
\label{tab:judge_ablation}
\setlength{\tabcolsep}{1mm}
\scalebox{0.9}{
\begin{tabular}{lccccccc}
\toprule
LLM Judge & Params & Acc & Input (¥/M) & Output (¥/M) & Input Tokens & Output Tokens & API Cost \\
\midrule
Qwen3.7-Max       & --      & 0.361 & 6  & 18 & 26.8M & 0.31M & ¥167 \\
DeepSeek-V4-Pro   & 1.6T    & 0.358 & 12 & 24 & 28.4M & 0.33M & ¥349 \\
DeepSeek-V4-Flash & 284B    & 0.351 & 1  & 2  & 31.6M & 0.39M & ¥32 \\
Qwen3.6-Flash     & 35B     & 0.338 & 1.2& 7.2& 34.7M & 0.43M & ¥45 \\
\bottomrule
\end{tabular}}
\end{table}

We further retrain SkillMentor with four different judges on Qwen3.5-4B. Table~\ref{tab:judge_ablation} reports the resulting performance and estimated API cost of 200 steps. Performance varies only modestly, from 0.338 (Qwen3.6-Flash) to 0.361 (Qwen3.7-Max), and even the weakest judge still substantially outperforms the No Skill baseline (0.244). In contrast, API cost differs by more than an order of magnitude. DeepSeek-V4-Flash achieves 0.351 accuracy at only ¥32, remaining within one point of the strongest judge (0.361) while costing roughly one-fifth as much.

Together, these results show that SkillMentor's reliance on strong models is both light and robust: the frontier model serves a single, deployment-free role, and its specific identity has little effect on the quality of learned diagnosis, eliminating the need for expensive flagship-tier models as even low-cost Flash-tier APIs suffice.

\section{Conclusion}

We study a previously overlooked component of agent self-evolution: \emph{diagnosis}. Rather than learning how to act, we ask whether an agent can learn to discover another agent's blind spots. To isolate this capability, we considered the cleanest setting in which the executor remains frozen and no human-labeled data is available.

We propose SkillMentor, which trains a small Mentor policy through reinforcement learning to jointly perform blind-spot discovery and skill curation. By externalizing corrective knowledge as an evolving skill repository, SkillMentor turns self-evolution into a diagnosis problem rather than a weight adaptation problem. Across AppWorld and BFCLv3, SkillMentor consistently improves frozen executors, transfers across models without retraining, and produces interpretable, reusable skills stored as human-readable Markdown files.

More broadly, our results suggest that blind spots are not static failures to be patched once, but explicit entities that can be discovered, repaired, transferred, and continually re-diagnosed as an executor evolves. This perspective also separates diagnosis from execution: a small RL-trained Mentor can outperform larger prompt-based models, indicating that diagnosis itself is a learnable capability rather than a byproduct of model scale.

We hope this work encourages future research on treating diagnosis as a fundamental component of self-evolving agents and on developing dedicated policies that learn \emph{what an agent does not know}, rather than solely improving \emph{how an agent acts}.

\bibliography{iclr2025_conference}
\bibliographystyle{iclr2025_conference}

\appendix

\section{Hyperparameters}
\label{app:config}

Table~\ref{tab:hyperparams} lists the key hyperparameters used in all experiments.

\begin{table}[h]
\centering
\caption{Key hyperparameters for SkillMentor.}
\label{tab:hyperparams}
\setlength{\tabcolsep}{2.7mm}
\scalebox{0.86}{
\begin{tabular}{lcl}
\toprule
Component & Parameter & Value \\
\midrule
\multirow{7}{*}{Training} & Total training steps & 200 \\
                          & Test frequency & 20 steps \\
                          & GRPO rollouts group $G$ & 8 \\
                          & Learning rate & $1 \times 10^{-6}$ \\
                          & Max prompt length & 12,000 tokens \\
                          & Max response length & 2,048 tokens \\
                          & KL coefficient $\beta$ & 0.01 \\
\midrule
\multirow{4}{*}{Mentor} & Model & Qwen3.5-4B \\
                          & Max prompt length & 10,240 tokens \\
                          & Max response length & 3,072 tokens \\
                          & Summarizer max prompt & 8,000 tokens \\
\midrule
\multirow{4}{*}{Blind-Spot Discovery} & Free exploration steps $K$ & 10 \\
                               & Candidate tasks $N$ & 8 \\
                               & Task diversity window $W$ & 8 \\
                               & Gap threshold $\tau$ & 0.5 (decaying) \\
\midrule
\multirow{4}{*}{Skill Curation} & Candidate skills $M$ & 8 \\
                               & Format weight $\alpha_f$ & 0.3 \\
                               & Quality weight $\alpha_q$ & 0.7 \\
                               & Format validation & Whitelist of App/API names \\
\midrule
\multirow{3}{*}{LLM Judge} & Model & DeepSeek-V4-Flash \\
                            & Executor (frozen) & Qwen3.5-0.6B to 9B \\
                            & vLLM GPU memory & 45\% \\
\midrule
Hardware & GPUs & 4$\times$NVIDIA A100 \\
\bottomrule
\end{tabular}}
\end{table}

\section{Hyperparameter Sensitivity}
\label{app:sensitivity}

We verify that performance is robust to key hyperparameter choices. Table~\ref{tab:sensitivity} reports AppWorld accuracy for the Qwen3.5-4B executor under three sweeps, each varying one parameter while holding the others at their defaults.

\begin{table}[h]
\centering
\caption{Hyperparameter sensitivity on AppWorld (Qwen3.5-4B executor). Default settings are marked with $\dagger$.}
\label{tab:sensitivity}
\setlength{\tabcolsep}{3mm}
\scalebox{0.88}{
\begin{tabular}{llcc}
\toprule
Parameter & Value & Acc & Step \\
\midrule
\multirow{3}{*}{Format weight $\alpha_f$}
  & 0.2 & 0.344 & 23.1 \\
  & 0.3 $\dagger$ & \textbf{0.351} & \textbf{22.7} \\
  & 0.4 & 0.347 & 23.0 \\
\midrule
\multirow{3}{*}{Gap threshold $\tau$}
  & 0.3 & 0.339 & 23.4 \\
  & 0.5 $\dagger$ & \textbf{0.351} & \textbf{22.7} \\
  & 0.7 & 0.332 & 23.8 \\
\midrule
\multirow{3}{*}{Eviction threshold}
  & 0.02 & 0.346 & 22.9 \\
  & 0.05 $\dagger$ & \textbf{0.351} & \textbf{22.7} \\
  & 0.10 & 0.342 & 23.2 \\
\bottomrule
\end{tabular}}
\end{table}

Performance is stable across all settings: the maximum drop is 1.9 points (gap threshold 0.7 vs.\ 0.5). The gap threshold $\tau$ is the most sensitive parameter since it controls which failures proceed to curation. Reward weighting and eviction threshold are less sensitive.

\section{Prompts}
\label{app:prompts}

We provide the full prompt templates used in the SkillMentor training loop. All prompts are reproduced verbatim from our implementation.

\begin{figure}[tp]
\centering
\includegraphics[width=\linewidth]{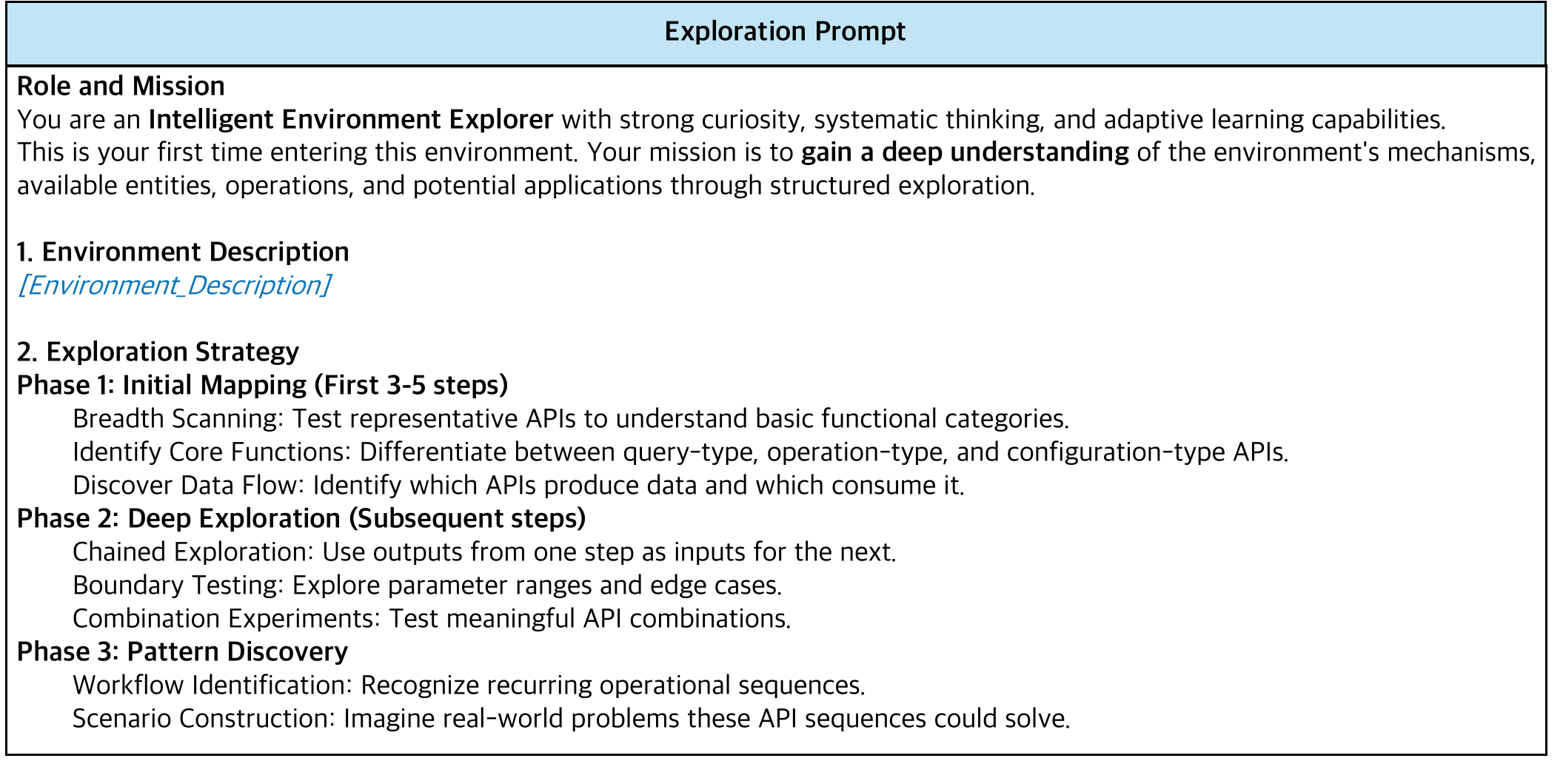}
\caption{Exploration prompt.}
\label{fig:prompt_explore}
\end{figure}

\textbf{Exploration Prompt} (Figure~\ref{fig:prompt_explore}) shows the system prompt for the Mentor's free exploration phase, instructing it to systematically map available tools and APIs.

\begin{figure}[tp]
\centering
\includegraphics[width=\linewidth]{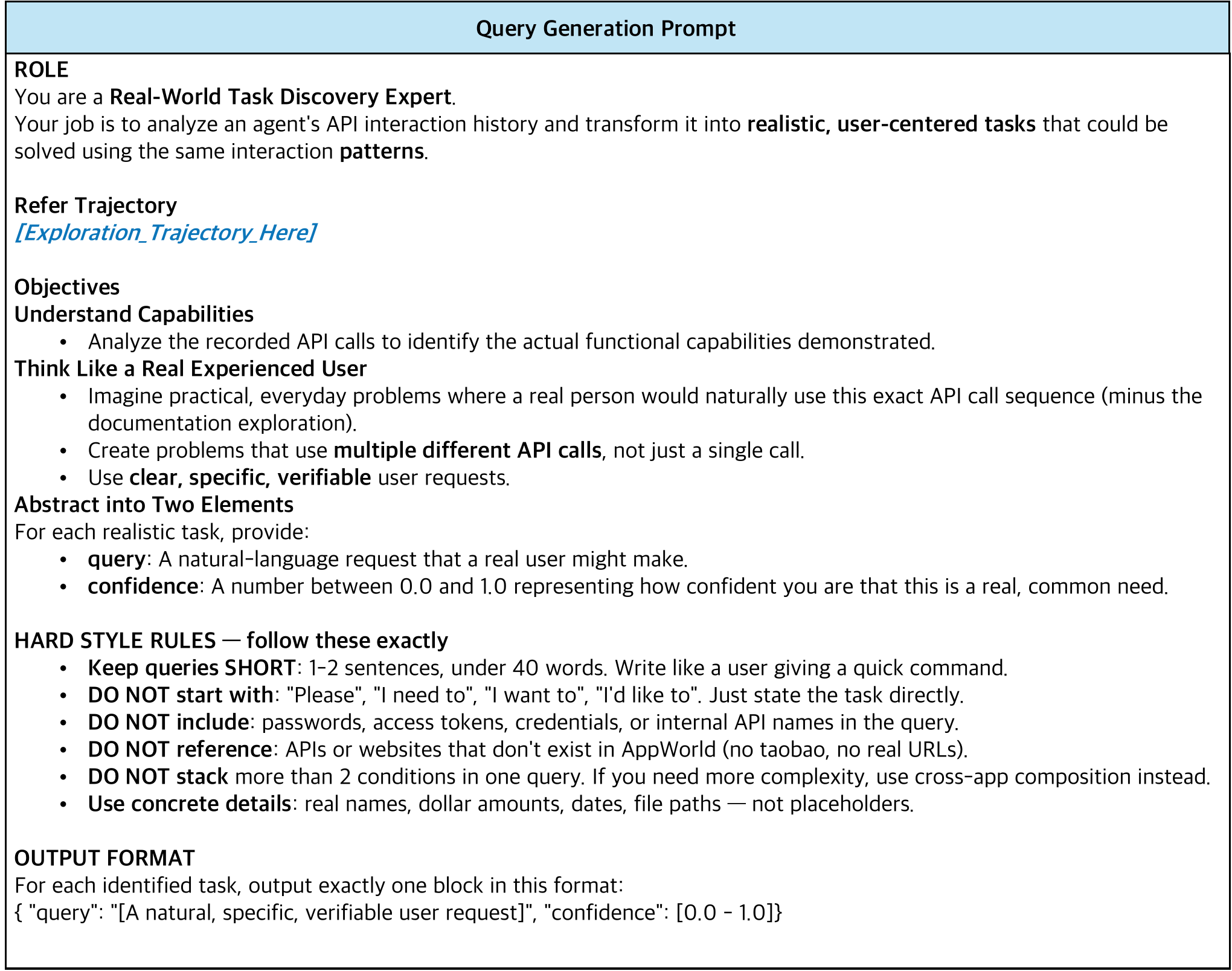}
\caption{Task generation prompt.}
\label{fig:prompt_question}
\end{figure}

\textbf{Task Generation Prompt} (Figure~\ref{fig:prompt_question}) specifies hard style rules, output format, and in-context examples for generating candidate diagnostic tasks.

\begin{figure}[tp]
\centering
\includegraphics[width=\linewidth]{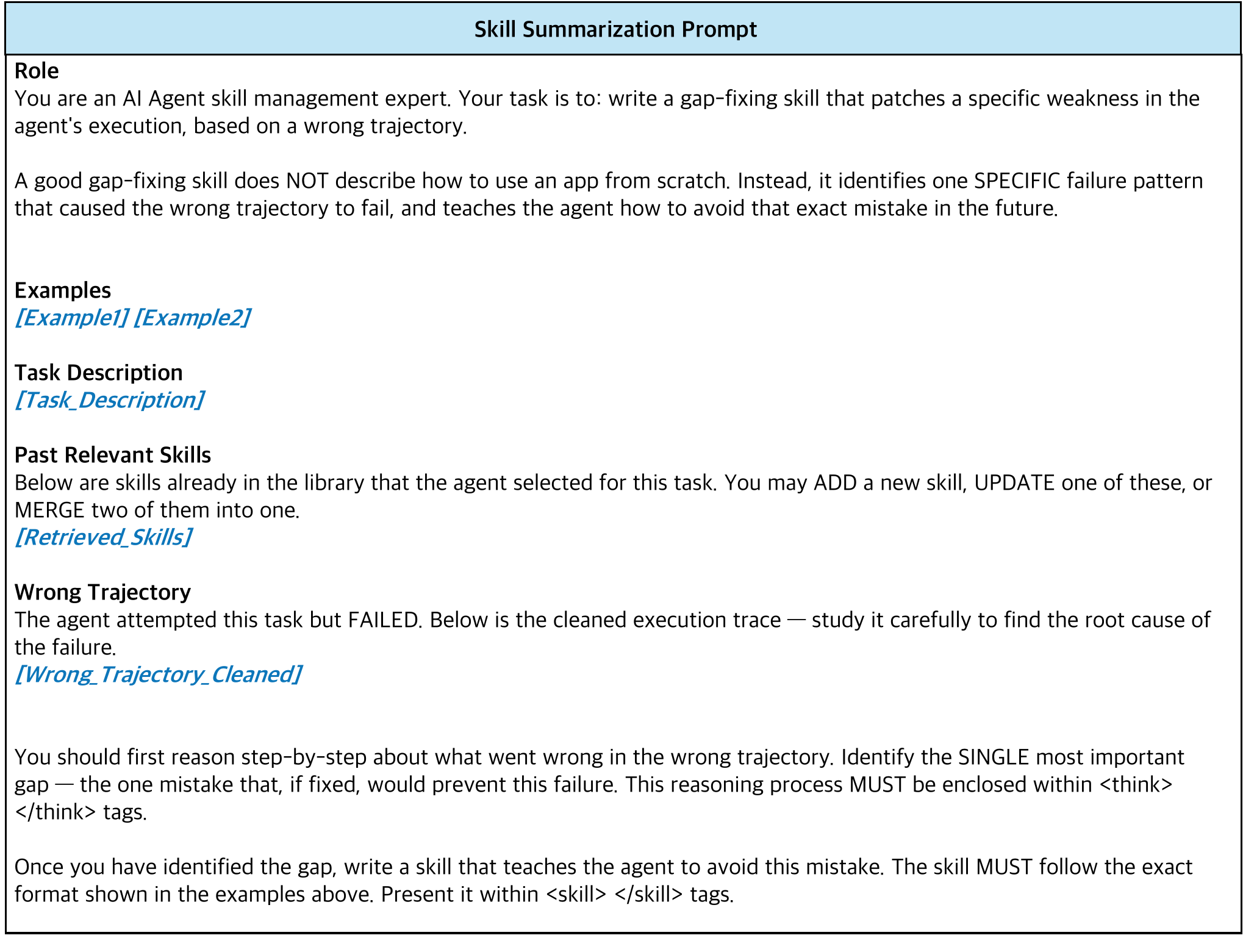}
\caption{Skill summarization prompt.}
\label{fig:prompt_summarize}
\end{figure}

\textbf{Skill Summarization Prompt} (Figure~\ref{fig:prompt_summarize}) defines the Markdown output format with YAML frontmatter, trigger conditions, step-by-step rules, and management actions (\texttt{ADD}, \texttt{UPDATE}, \texttt{MERGE}).

\begin{figure}[tp]
\centering
\includegraphics[width=\linewidth]{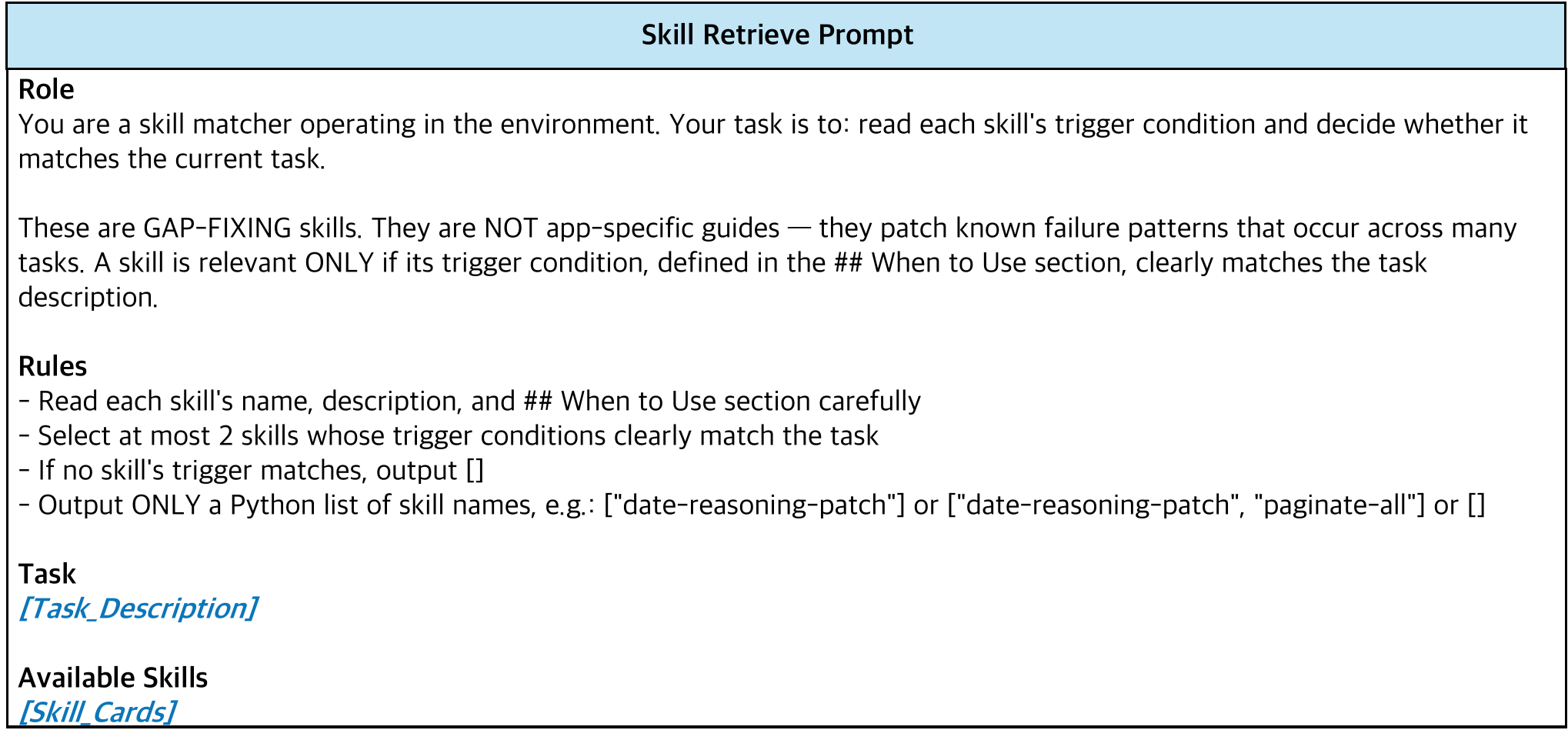}
\caption{Skill retrieval prompt.}
\label{fig:prompt_retrieve}
\end{figure}

\textbf{Skill Retrieval Prompt} (Figure~\ref{fig:prompt_retrieve}) uses a structured scoring rubric across multiple relevance dimensions.

\begin{figure}[tp]
\centering
\includegraphics[width=\linewidth]{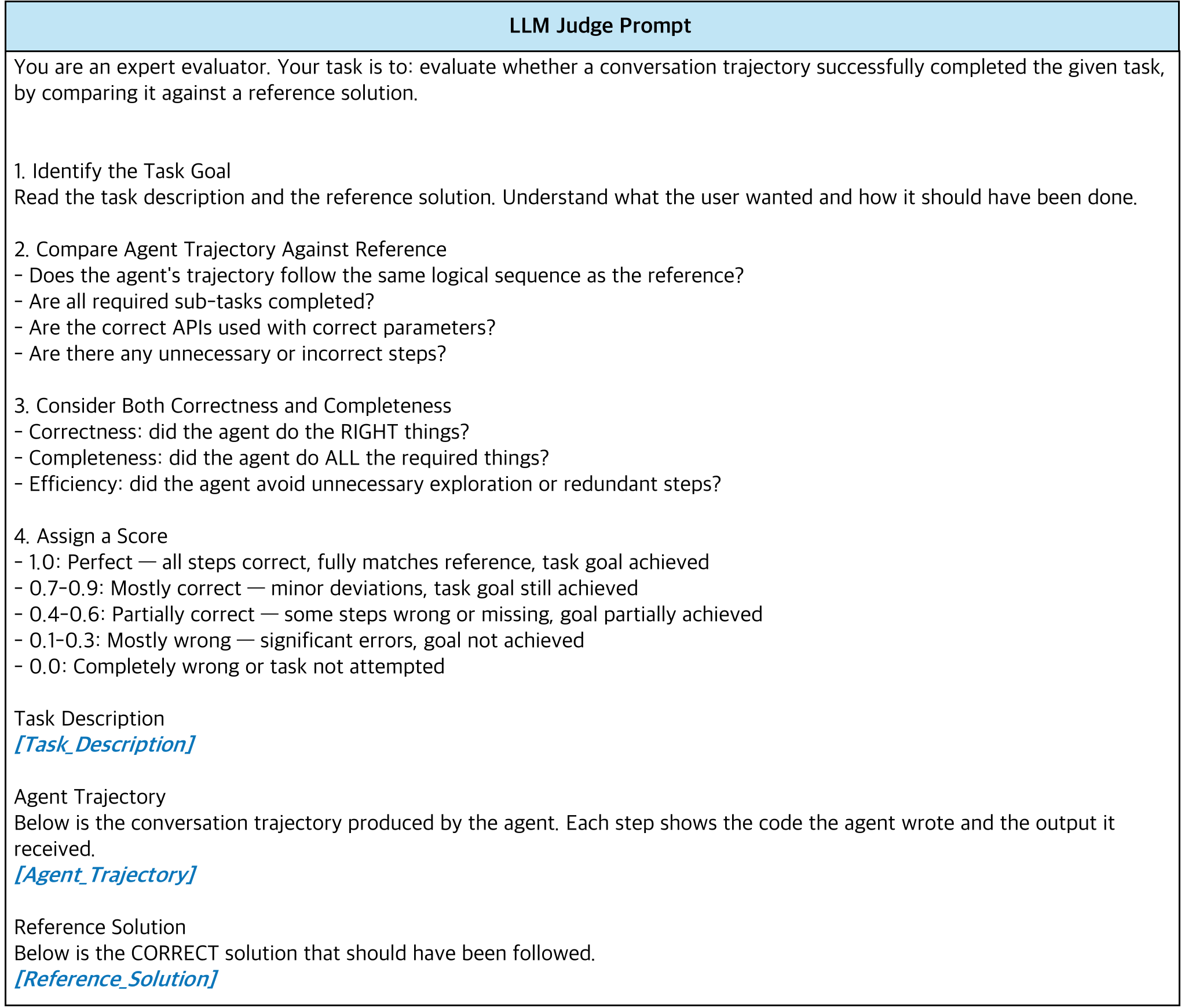}
\caption{LLM judge prompt.}
\label{fig:prompt_judge}
\end{figure}

\textbf{LLM Judge Prompt} (Figure~\ref{fig:prompt_judge}) defines four evaluation dimensions (goal achievement, action correctness, efficiency, completeness) and a detailed scoring rubric.

\section{Qualitative Skill Examples}
\label{app:skill_examples}

\begin{figure}[tp]
\centering
\includegraphics[width=\linewidth]{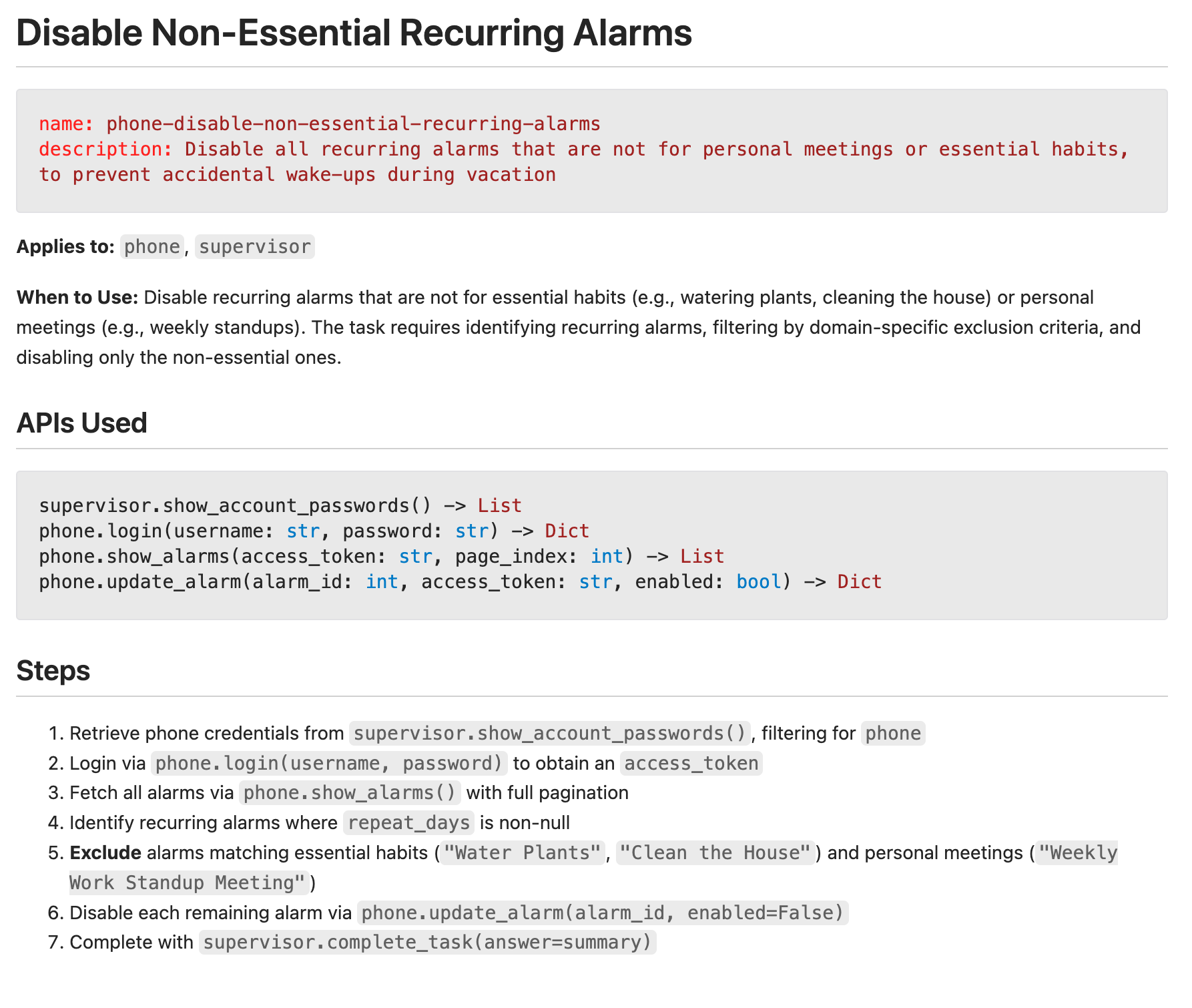}
\caption{\texttt{spotify-liked-songs-from-followed-artists-with-reviews}.}
\label{fig:skill_examples_a}
\end{figure}

\begin{figure}[tp]
\centering
\includegraphics[width=\linewidth]{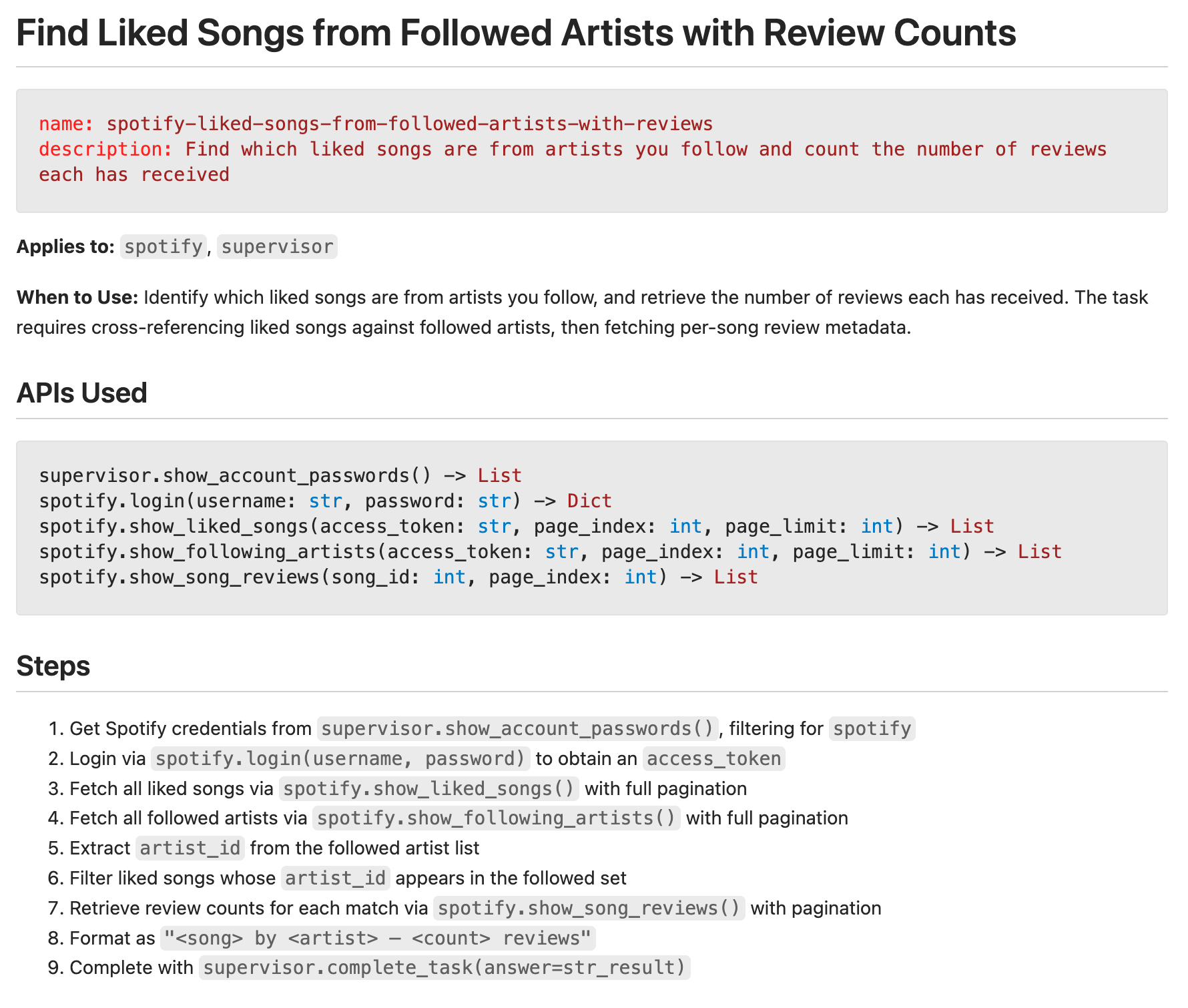}
\caption{\texttt{phone-disable-non-essential-recurring-alarms}.}
\label{fig:skill_examples_b}
\end{figure}

Figures~\ref{fig:skill_examples_a} and~\ref{fig:skill_examples_b} present two representative skills from the final library. The \texttt{spotify-liked-songs-from-followed-artists-with-reviews} encodes a nine-step procedure that cross-references two paginated datasets: liked songs and followed artists must each be fully loaded via pagination before computing their intersection in memory, after which per-song review counts are retrieved. The executor consistently failed by joining the datasets prematurely or skipping the review-count retrieval step. The \texttt{phone-disable-non-essential-recurring-alarms} encodes an exclusion-based filtering logic: rather than enumerating which alarms to disable, it identifies recurring alarms by the \texttt{repeat\_days} field and then excludes only those matching a small set of essential patterns (watering plants, cleaning, weekly standup). This negative-filter design makes the skill robust to changes in the alarm list. Each skill follows a structure the Mentor learned to produce through RL: a YAML frontmatter, a trigger condition, step-by-step rules, and a common-mistakes section.

\section{Skill Rescue Case Studies}
\label{app:rescued}

A direct way to measure the value of a curated skill is to observe whether injecting it into the executor's context rescues a previously failed task. Table~\ref{tab:rescued} reports five cases drawn from the validation set. In each case, the executor (Qwen3.5-4B) was first run without any skill repository and produced an incorrect or incomplete result (score 0.0). The same executor was then re-run on the identical task, this time with a single relevant skill from the final library prepended to its system prompt. All five tasks were rescued (score 0$\to$1).

\begin{table}[tp]
\centering
\caption{Skill rescue cases. Each failed without skills and succeeded after a single relevant skill was injected.}
\label{tab:rescued}
\setlength{\tabcolsep}{1mm}{
\scalebox{0.83}{
\renewcommand{\cellalign}{tl}
\begin{tabular}{lllll}
\hline
Task & Skill & Score & Steps & Root Cause \\
\hline
\makecell{Import markdown files \\ to Simple Note} & \texttt{file\_system-auth-param-patch} & 0$\to$1 & 39$\to$25 & \makecell{Missing \texttt{access\_token}, \\ wrong param name} \\
\makecell{Find recipe in Simple Note, \\ reply on phone} & \texttt{find-existing-note-before-creating} & 0$\to$1 & 39$\to$43 & \makecell{Creating new note \\ instead of searching} \\
\makecell{Reorganize meeting files \\ by date prefix} & \texttt{file\_system-auth-param-patch} & 0$\to$1 & 33$\to$33 & \makecell{Missing \texttt{access\_token} \\ in API call} \\
\makecell{Accept Venmo \\ carpool request} & \texttt{venmo-pending-request-patch} & 0$\to$1 & 14$\to$39 & \makecell{Queried past transactions \\ instead of pending requests} \\
\makecell{Reset Spotify queue, \\ shuffle, and play} & \texttt{spotify-artist-following-patch} & 0$\to$1 & 25$\to$33 & \makecell{Using wrong API \\ (\texttt{clear\_queue})} \\
\hline
\end{tabular}}}
\end{table}

Two patterns stand out. First, the failures are mechanical rather than strategic: the executor knows what to do but stumbles on API syntax, parameter names, or authentication flow. Second, the skills that fix these failures are concise (each under 25 lines) and encode a single procedural rule. This is precisely the type of knowledge that SkillMentor's format validation and executor-grounded quality assessment are designed to capture.

\begin{figure}[tp]
\centering
\includegraphics[width=\textwidth]{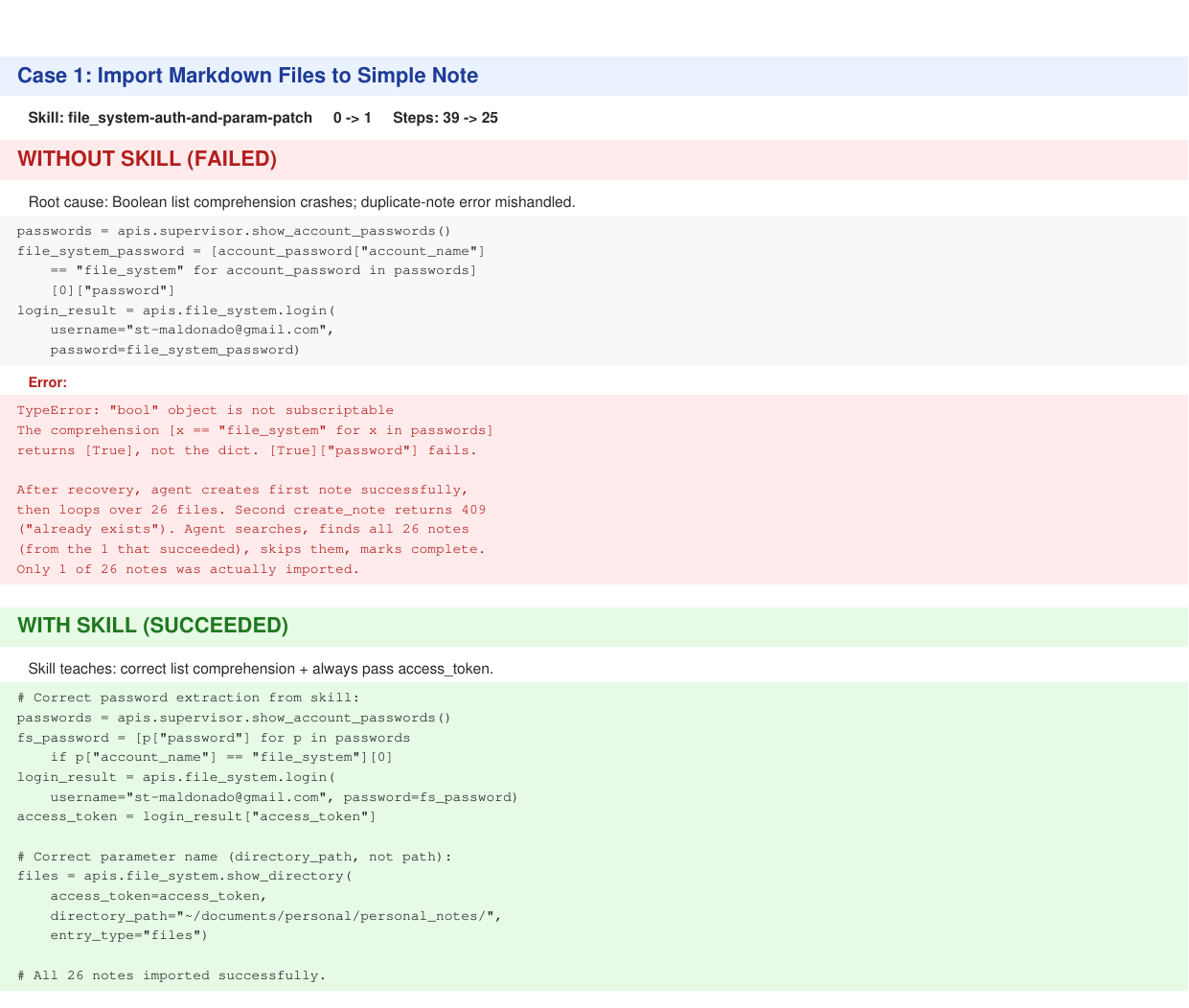}
\caption{\textbf{Case 1:} Import markdown files to Simple Note. The executor crashes on a boolean list comprehension when extracting the password (red), then mishandles a duplicate-note error. With the skill \texttt{file\_system-auth-and-param-patch} (green), it uses the correct comprehension pattern and completes all 26 imports.}
\label{fig:case1}
\end{figure}

\begin{figure}[tp]
\centering
\includegraphics[width=\textwidth]{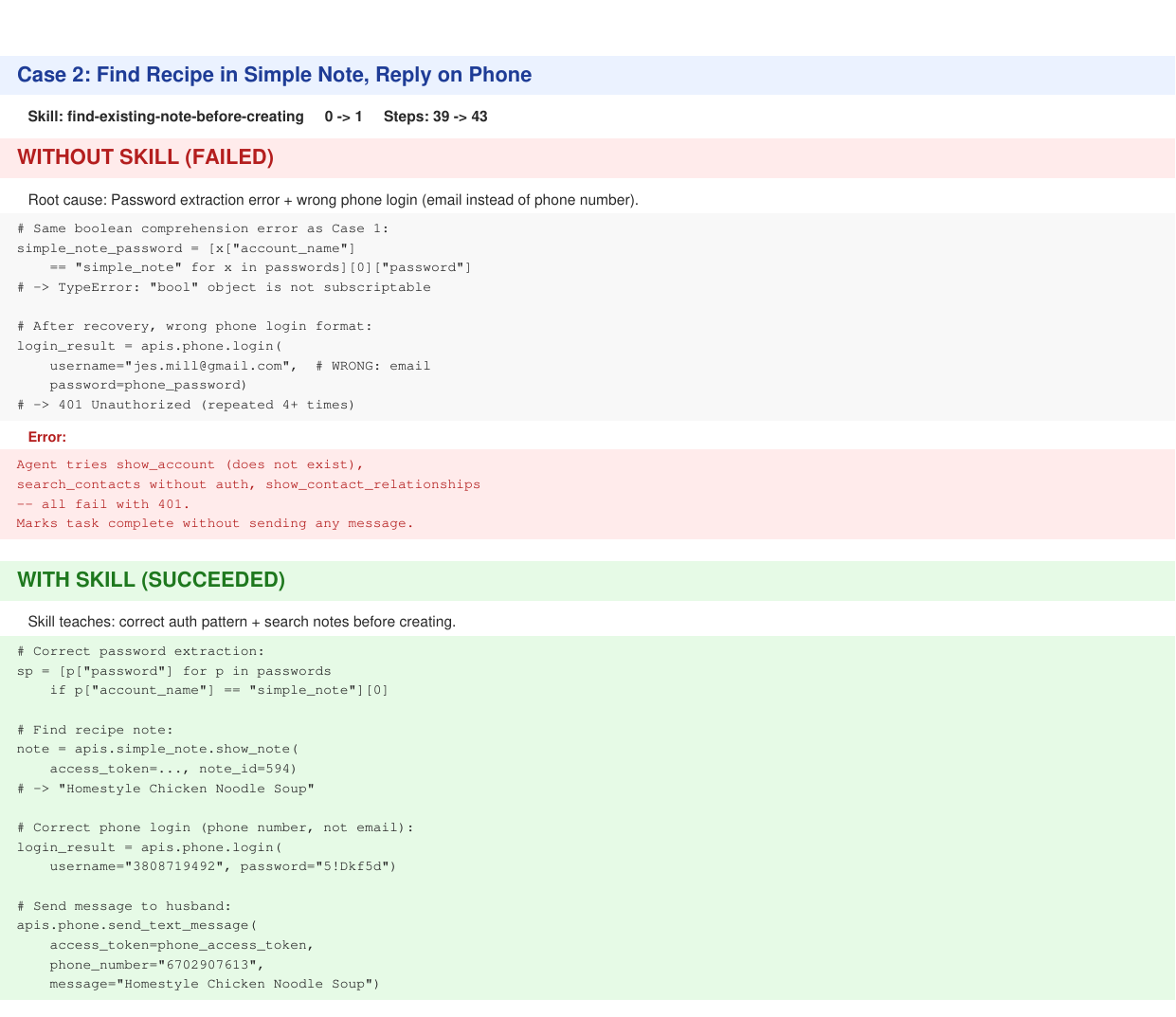}
\caption{\textbf{Case 2:} Find recipe in Simple Note, reply on phone. Without the skill, the executor hits the same boolean comprehension error and then uses email instead of phone number for phone login, causing repeated 401 failures. With \texttt{find-existing-note-before-creating}, it extracts the recipe correctly and logs into phone with the phone number, sending the message successfully.}
\label{fig:case2}
\end{figure}

\begin{figure}[tp]
\centering
\includegraphics[width=\textwidth]{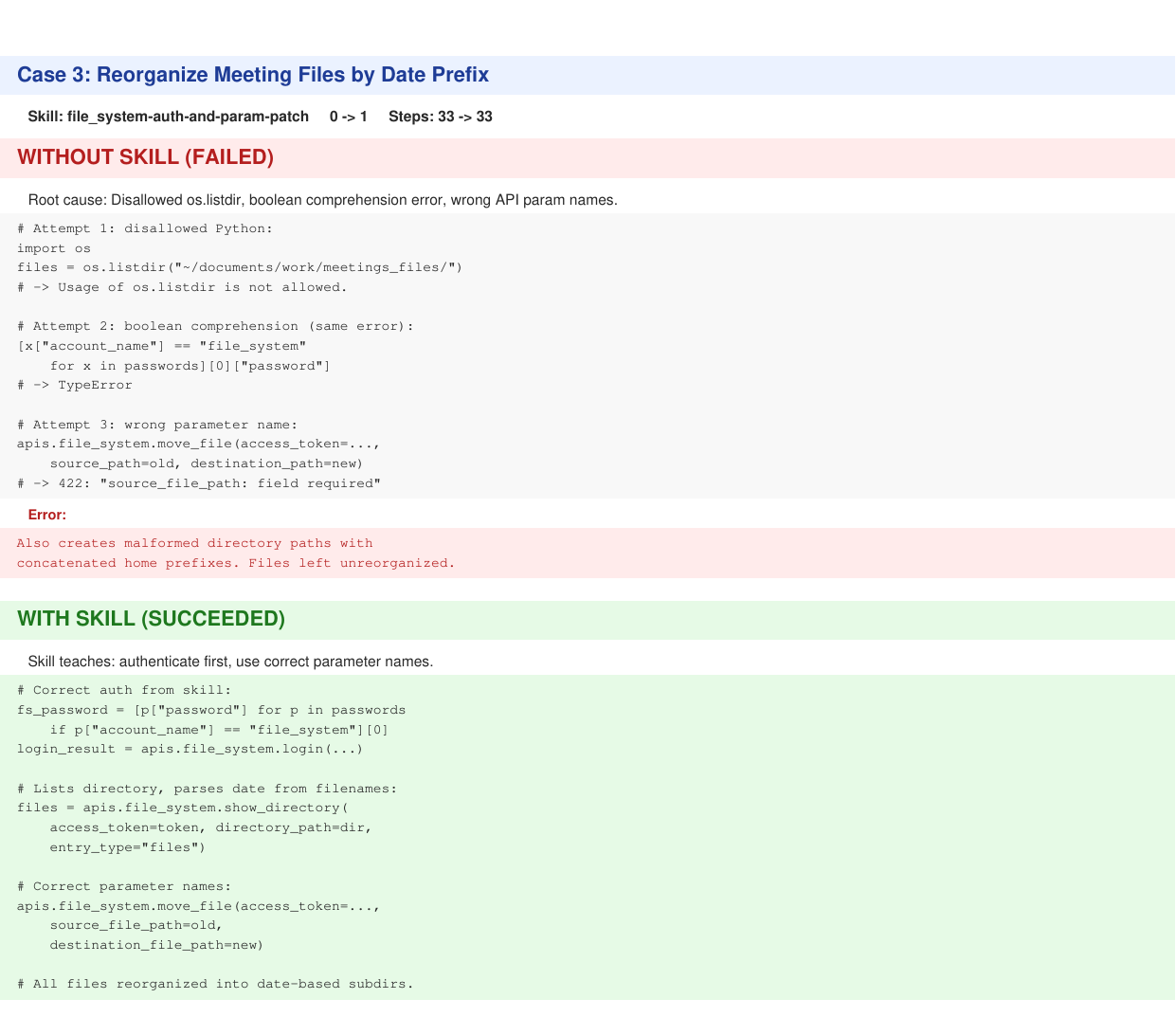}
\caption{\textbf{Case 3:} Reorganize meeting files by date prefix. The executor starts with disallowed \texttt{os.listdir}, then the boolean comprehension error, then uses wrong parameter names (\texttt{source\_path} vs. \texttt{source\_file\_path}). With \texttt{file\_system-auth-and-param-patch}, it authenticates correctly and uses proper parameter names to reorganize all files.}
\label{fig:case3}
\end{figure}

\begin{figure}[tp]
\centering
\includegraphics[width=\textwidth]{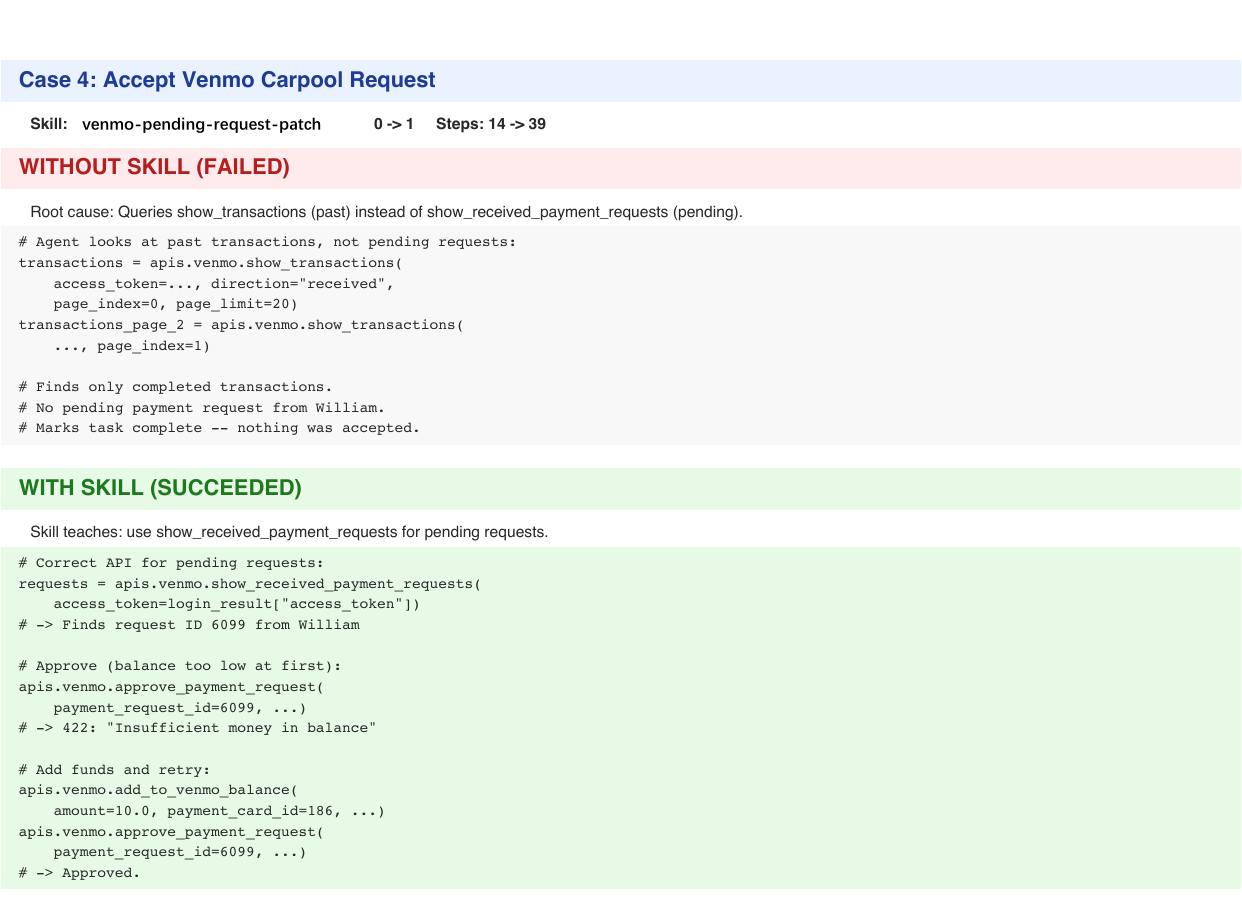}
\caption{\textbf{Case 4:} Accept Venmo carpool request. The executor queries \texttt{show\_transactions} (past transactions) instead of \texttt{show\_received\_payment\_requests} (pending requests), and marks the task complete without accepting anything. With \texttt{venmo-pending-request-patch}, it finds the pending request, tops up the balance, and approves it.}
\label{fig:case4}
\end{figure}

\begin{figure}[tp]
\centering
\includegraphics[width=\textwidth]{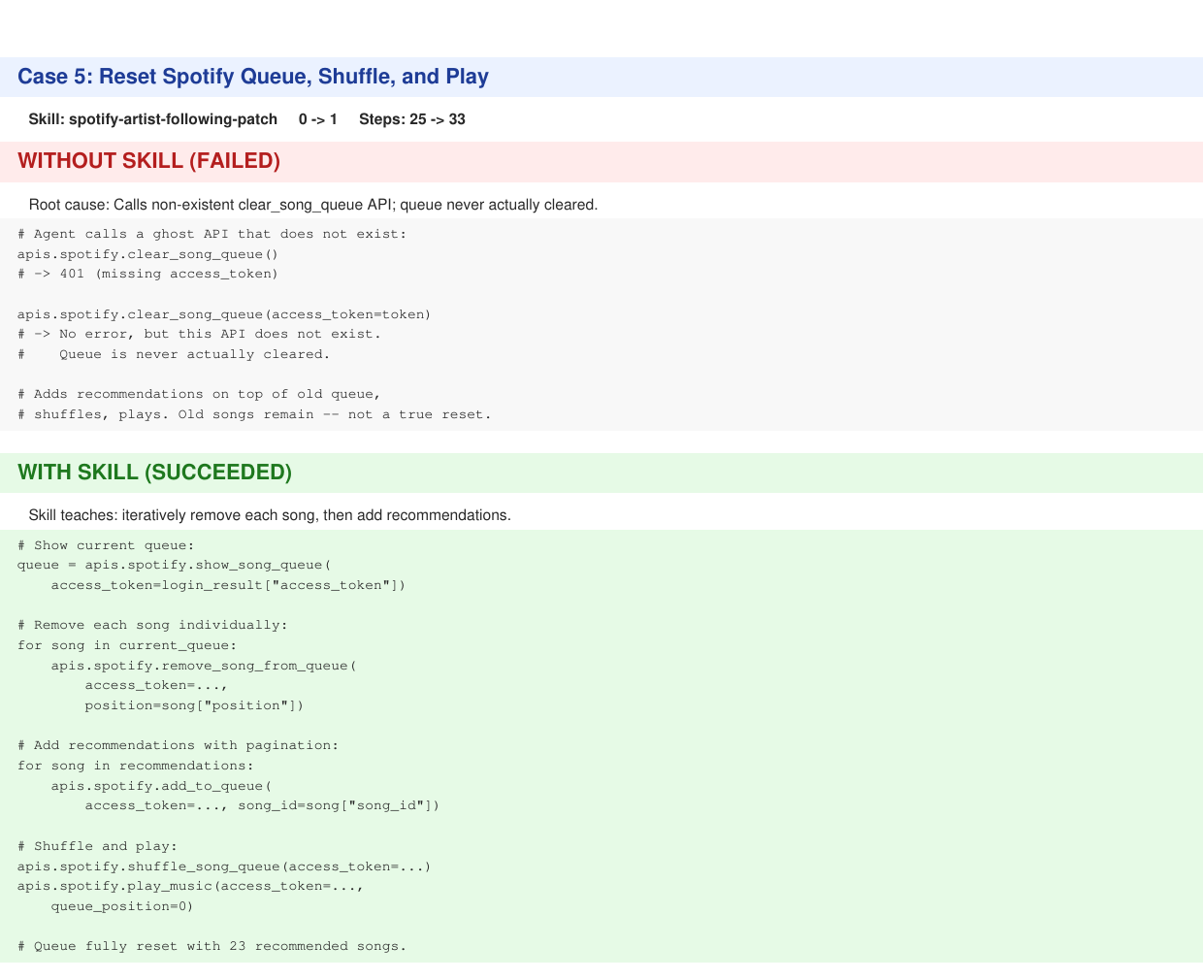}
\caption{\textbf{Case 5:} Reset Spotify queue, shuffle, and play. The executor calls a non-existent \texttt{clear\_song\_queue} API; the queue is never actually emptied. With \texttt{spotify-artist-following-patch}, it iteratively removes each song via \texttt{remove\_song\_from\_queue}, then repopulates and shuffles.}
\label{fig:case5}
\end{figure}

Figures~\ref{fig:case1}--\ref{fig:case5} present the full trajectory comparison for each rescue case, showing the critical failure point in the unassisted execution (red) and how the injected skill guides the executor to a successful outcome (green).

\end{document}